\DocumentMetadata{}

\let\checkmark\relax

\documentclass[sigconf, nonacm]{acmart}
\setcopyright{none}
\usepackage{xcolor}
\usepackage[normalem]{ulem}
\usepackage{url}
\usepackage{graphicx}
\usepackage{enumitem}
\usepackage{pifont, tabularx, booktabs}
\usepackage{tablefootnote}
\usepackage{multirow}
\usepackage[dvipsnames]{xcolor}
\usepackage{colortbl}
\usepackage{tikz}

\usetikzlibrary{patterns, pgfplots.groupplots}

\usepackage{pgfplots}

\usepackage{caption}
\usepackage{subcaption}
\usepackage{listings}
\usepackage{upquote}
\usepackage{tcolorbox}
\usepackage{fontawesome5}
\usepackage{siunitx}
\usepackage{textcase}

\usepackage{pifont, tabularx, booktabs}
\newcommand{\xmark}{\ding{55}} 

\usepackage[dvipsnames]{xcolor}
\definecolor{amber}{RGB}{255,191,0}

\setlist[itemize]{leftmargin=*, nosep}

\definecolor{clrDirect}{RGB}{30,144,255}
\definecolor{clrLangG}{RGB}{46,204,113}
\definecolor{clrCrew}{RGB}{231,76,60}
\definecolor{myred}{RGB}{255,0,0}
\definecolor{mygreen}{RGB}{0,128,0}

\newcommand{\MAFBench}{\mbox{MAFBench}}

\newcommand{\gradientcell}[1]{#1}

\newcommand{\accChange}[2]{%
\pgfmathsetmacro{\diff}{#2 - #1}%
\pgfmathsetmacro{\len}{max(.1, min(0.05*abs(\diff), .3))}%
\ifdim \diff pt > 0pt #2\, \tikz[x=1mm, >=stealth, line width=0.2mm]{ \draw[->, green!70!black] (1,0) -- ++(0,{0.8*\len}); }%
\else #2\, \tikz[x=1mm, >=stealth, line width=0.2mm]{ \draw[->, red!70!black] (1,0) -- ++(0,{-0.8*\len}); }%
\fi }

\usepackage{xcolor}
\definecolor{revonetext}{HTML}{185FA5} 
\definecolor{revonebg}{HTML}{F0E9FA} 
\definecolor{revonefg}{HTML}{5B2E9D} 
\definecolor{revtwotext}{HTML}{0F6E56} 
\definecolor{revtwobg}{HTML}{E1F5EE}
\definecolor{revtwofg}{HTML}{085041}
\definecolor{revthreetext}{HTML}{993C1D} 
\definecolor{revthreebg}{HTML}{FAECE7}
\definecolor{revthreefg}{HTML}{712B13}

\colorlet{revbodytext}{revonetext} 

\newif\ifshowrev
\newif\ifunifiedrevtext

\ifshowrev
\ifunifiedrevtext
\colorlet{revonebody}{revbodytext}
\colorlet{revtwobody}{revbodytext}
\colorlet{revthreebody}{revbodytext}
\else
\colorlet{revonebody}{revonetext}
\colorlet{revtwobody}{revtwotext}
\colorlet{revthreebody}{revthreetext}
\fi
\newcommand{\revone}[2]{\textcolor{revonebody}{#2}~\revbadge{revonebg}{revonefg}{#1}}
\newcommand{\revtwo}[2]{\textcolor{revtwobody}{#2}~\revbadge{revtwobg}{revtwofg}{#1}}
\newcommand{\revthree}[2]{\textcolor{revthreebody}{#2}~\revbadge{revthreebg}{revthreefg}{#1}}
\else
\newcommand{\revone}[2]{#2}
\newcommand{\revtwo}[2]{#2}
\newcommand{\revthree}[2]{#2}
\fi

\newcommand{\todo}[1]{}

\def\shorten{\looseness=-1}

\acmYear{2027}

\title[Architectural Design, Not Only Model Intelligence, Governs Multi-Agent LLM Performance]{%
Architectural Design, Not Only Model Intelligence, Governs Multi-Agent LLM Performance 
}

\author{Abdelghny Orogat}
\affiliation{ \institution{Concordia University} \country{}}

\author{Ana Rostam}
\affiliation{\institution{Concordia University}\country{}}

\author{Essam Mansour}
\affiliation{\institution{Concordia University}\country{}}

\begin{document}
  \sloppy

  \begin{abstract}
  Multi-agent LLM frameworks are data-intensive systems that govern how agents orchestrate tasks, manage state, and coordinate decisions. These architectural choices control execution overhead, memory behavior, planning effectiveness, and coordination scalability. Their impact on system performance remains poorly understood. Existing benchmarks evaluate individual agent capabilities in isolation and lack standardized framework-level comparison. We make four contributions. We introduce an architectural taxonomy that decomposes multi-agent LLM frameworks along five dimensions: orchestration, memory, planning interfaces, specialization, and communication topology. We develop \MAFBench{}, a unified evaluation suite that integrates existing benchmarks within a standardized execution pipeline. We conduct a controlled empirical study across nine frameworks, fixing the underlying LLM and varying only architectural design choices. We distill the results into six evidence-based design principles. Architectural design, not only model intelligence, governs performance. Orchestration alone increases latency by over $60\times$, and a minimal implementation of the same paradigm isolates that cost as implementation rather than paradigm. Schema-constrained planning interfaces reduce accuracy by up to 32 points through formatting failures, not reasoning errors. Communication topology drops coordination success from above $90\%$ to below $30\%$ under mismatched structure. Memory architecture controls recall and scalability independent of context window size, and no evaluated framework natively supports controlled knowledge revision.
\end{abstract}

  \maketitle

  \vspace{-1mm}
\section{Introduction}
\label{sec:introduction}

\definecolor{BestGreen}{RGB}{0,120,0}
\definecolor{WorstRed}{RGB}{160,0,0}
\setlength{\tabcolsep}{3.5pt}
\begin{table}[t]
\centering
\renewcommand{\arraystretch}{0.8}
\caption{Performance ranges across architectural design choices at fixed task
and LLM. \revone{R1O2}{Residual is framework time outside the model call, and it
separates paradigm cost from implementation cost.}}
\vspace{-3mm}
\label{tab:motivation}
\begin{tabular}{@{}llrr@{}}
\toprule
\textbf{Dimension} & \textbf{Metric} & \textbf{Best} & \textbf{Worst} \\
\midrule
Orchestration & Latency ($\times$ direct LLM) & {\color{BestGreen}$1.1\times$} & {\color{WorstRed}$62\times$} \\
              & Throughput (req/s)            & {\color{BestGreen}$7.0$}      & {\color{WorstRed}$0.12$} \\
              & \revone{R1O2}{Residual (\% latency)} & {\color{BestGreen}$<0.1$} & {\color{WorstRed}$91.5$} \\
\addlinespace[1pt]
Memory        & Overall score (\%)            & {\color{BestGreen}$23.8$}     & {\color{WorstRed}$6.1$} \\
\addlinespace[1pt]
Planning      & Accuracy (pts vs.\ no plan)   & {\color{BestGreen}$+15.7$}    & {\color{WorstRed}$-32.3$} \\
              & Runtime ($\times$ no plan)    & {\color{BestGreen}$0.8\times$}& {\color{WorstRed}$31.3\times$} \\
\addlinespace[1pt]
Specialization& $\Delta$F1 (vs.\ no role)     & {\color{BestGreen}$+58$}      & {\color{WorstRed}$\approx 0$} \\
\addlinespace[1pt]
Topology      & Success, large $n$ (\%)       & {\color{BestGreen}$>90$}      & {\color{WorstRed}$<30$} \\
\bottomrule
\end{tabular}
\end{table} 

\begin{figure}[t]
  \centering
  \includegraphics[width=.98\linewidth, trim=0.0cm 5.3cm 11.4cm 2.8cm, clip]{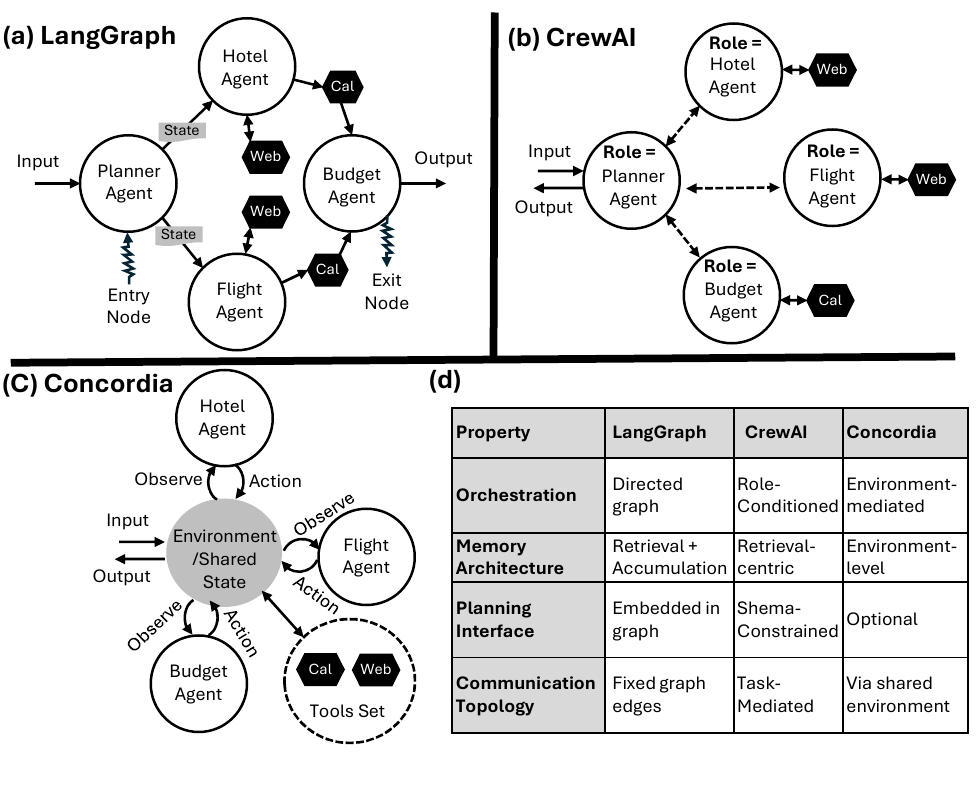}
  \vspace{-3mm}
  \caption{Running example: the same task (trip planning) instantiated under three paradigms (a--c). Panel (d) compares their design choices along the architectural dimensions evaluated in this study. \textit{Web (Search)/Cal(Calculator)} denote tools.}
  \label{fig:running_example}
\end{figure}

Multi-agent LLM frameworks govern how agents orchestrate tasks, manage state, and coordinate decisions in systems powered by large language models (LLMs)~\cite{NEURIPS2024_fa54b0ed, proagentcooperativeagent, githubissuemultiagentllm, qian-etal-2024-chatdev, jin-etal-2024-agentreview}. A growing ecosystem of such frameworks has emerged to support complex intelligent applications~\cite{omar2025chatty, he2025llm, langgraph2025, crewai2025, wu2024autogen, openaiagents2025, vezhnevets2023generative, touzel2024simulation}. These frameworks function as data-intensive systems. Their architectural choices determine orchestration structure, storage design, planning interfaces, and communication topology. These choices shape system behavior much like query processing pipelines, storage engines, and distributed coordination protocols shape database performance.

This paper evaluates the framework orchestration layer, not the applications built on top of it. Their impact on system performance remains poorly understood. We evaluate five architectural dimensions: orchestration structure, memory architecture, planning interfaces, specialization conditioning, and communication topology. Table~\ref{tab:motivation} reports the performance ranges across these dimensions at fixed LLM and task. Architectural design alone induces order-of-magnitude differences in latency and throughput, and substantial variation in accuracy and task success. System behavior is therefore governed primarily by architectural design, not LLM quality alone.\shorten

To illustrate these differences, we consider a simple task: planning a travel itinerary with flights, accommodation, and a budget constraint. \revthree{R3W2}{Figure~\ref{fig:running_example} instantiates it under the three general-purpose paradigms of Table~\ref{tab:github-stats-updated}. Task-specific frameworks fix orchestration to one task family, so we treat them separately (Section~\ref{sec:background}).} LangGraph~\cite{langgraph2025} encodes orchestration as an explicit workflow graph. CrewAI~\cite{crewai2025} organizes agents through role-based delegation. Concordia~\cite{vezhnevets2023generative} mediates interaction through a shared environment. \revone{R1O2}{Concordia is the only production GABM framework, so we add a second data point: a \emph{minimal GABM skeleton} with the same environment-mediated structure but no simulation machinery. This makes two gaps measurable. Skeleton versus Concordia isolates implementation cost. Skeleton versus the other paradigms isolates the cost of environment mediation.} The task, agents, and LLM stay fixed, so each paradigm differs only in execution structure.\shorten

\begin{table}[t]
\vspace{-1mm}
\centering
\scriptsize
\setlength{\tabcolsep}{3.5pt}
\renewcommand{\arraystretch}{0.62}
\caption{Representative frameworks by paradigm, with the entity holding
control-flow authority in \emph{italics}. $\bullet$ marks frameworks evaluated
here. 
\revone{R1O2}{The skeleton gives the GABM paradigm a second data point.}}
\vspace{-3mm}
\label{tab:github-stats-updated}
\begin{tabular}{@{}p{1.25cm}p{2.8cm}rrrc@{}}
\toprule
\textbf{Paradigm} & \textbf{Framework} & \textbf{Stars} & \textbf{Forks} & \textbf{Contrib.} & \textbf{Eval.} \\
\midrule
\multirow{3}{*}{\shortstack[l]{Graph\\ \emph{(developer)}}}
& LangGraph~\cite{langgraph2025}   & 18.9k & 3.3k & 249  & $\bullet$ \\
& n8n~\cite{n8n2025}               & 140k  & 44k  & 400+ & \\
& Langflow~\cite{langflow2025}     & 19.6k & 2.1k & 120  & \\
\addlinespace[2pt]
\multirow{5}{*}{\shortstack[l]{Role\\ \emph{(agents)}}}
& CrewAI~\cite{crewai2025}         & 38.4k & 5.1k & 274  & $\bullet$ \\
& AutoGen~\cite{wu2024autogen}     & 50.0k & 7.7k & 558  & $\bullet$ \\
& OpenAI SDK~\cite{openaiagents2025} & 14.9k & 2.5k & 166 & $\bullet$ \\
& Agno~\cite{agno2025}             & 4.2k  & 310  & 40   & $\bullet$ \\
& OpenAgents~\cite{OpenAgents}     & 1.1k  & 143  & 5    & $\bullet$ \\
\addlinespace[2pt]
\multirow{2}{*}{\shortstack[l]{GABM\\ \emph{(environment)}}}
& Concordia~\cite{vezhnevets2023generative} & 1.0k & 218 & $<$50 & $\bullet$ \\
& \revone{R1O2}{GABM skeleton (ours)} & --- & --- & --- & $\bullet$ \\
\addlinespace[2pt]
\multirow{5}{*}{\shortstack[l]{Task-specific\\ \emph{(fixed)}}}
& \revthree{R3W2}{Claude Code~\cite{claudecode2026}} & 142k   & 22k   & 57  & \\
& \revthree{R3W2}{Cursor~\cite{cursor2026}}          & 33k   & 2.3k   & 34  & \\
& \revthree{R3W2}{Cline~\cite{cline2026}}            & 65.4k & 7.0k  & 334  & \\
& \revthree{R3W2}{OpenHands~\cite{openhands2025}}    & 82.2k & 10.7k & 538  & $\bullet$ \\
& \revthree{R3W2}{OpenClaw~\cite{openclaw2026}}      & 384k  & 80.7k & 1.6k & \\
\bottomrule
\end{tabular}
\vspace{-5mm}
\end{table}

Prior work examines individual aspects of multi-agent intelligence: planning~\cite{cobbe2021training, talmor-etal-2019-commonsenseqa, hendrycks2measuring}, memory and retrieval~\cite{hu2025memoryagentbench, hsieh2024ruler, li2018towards, xu2024detectiveqa, zhong2023mquake}, tool use~\cite{mialon2023gaiabenchmarkgeneralai, nguyen2024dynasaur, guo2024stabletoolbench, shi2025retrievalmodelsarenttoolsavvy}, and collaborative reasoning~\cite{coordinationcollaborativereasoning}. These studies target model-level capabilities, not framework-level architectural choices. Three limitations persist: (i)~no characterization of architectural dimensions and their role in system behavior, (ii)~no unified suite for cross-framework comparison, and (iii)~no controlled evidence isolating individual design choices.

We address these limitations with four contributions. Our study design is controlled. We fix the underlying LLM and vary only architectural design choices, so observed differences follow from architecture and not from model capability. Planning is evaluated across ten LLMs because the interface effect interacts with capability. All other dimensions hold the model constant. \revone{R1O2}{One paradigm ships with a single production framework, so we implement a minimal reference version of it to separate paradigm cost from implementation cost.} Our main contributions are:
\begin{itemize}
  \item An architectural taxonomy that compares multi-agent LLM frameworks along five dimensions: orchestration, memory architecture, planning interfaces, specialization conditioning, and communication topology. The taxonomy \revthree{R3M1}{raises five research questions that guide our evaluation} (Section~\ref{sec:taxonomy_analysis}).

  \item \textbf{MAFBench}\footnote{\faGithub\ \url{https://github.com/CoDS-GCS/MAFBench}}, a standardized evaluation suite that integrates existing benchmarks within a unified execution and scoring pipeline for controlled framework-level comparison (Section~\ref{sec:benchmark_collection}).

  \item A controlled empirical study across nine frameworks and five architectural dimensions. It shows that architectural design, not model intelligence alone, governs performance, scalability, and cost (Section~\ref{sec:eval}).

  \item Six evidence-based design principles that guide the construction and selection of multi-agent LLM frameworks. Each principle states where it applies and where it does not (Section~\ref{sec:design_principles}).
\end{itemize}

  \section{Architectural Taxonomy}
\label{sec:taxonomy_analysis} This section introduces our architectural taxonomy for multi-agent LLM frameworks. Architectural design choices shape control flow, agent abstraction, communication, memory, and execution semantics. They have first-order impact on performance, robustness, and scalability. The taxonomy enables framework-level comparison independent of specific models or tasks. \revone{R1O1}{We instantiate the trip-planning example of Figure~\ref{fig:running_example} in each paradigm and report its execution trace alongside the corresponding definition.} We begin with formal definitions.

\setlength{\tabcolsep}{2pt}  

\begin{table*}[t]
\vspace{-1mm}
\centering
\scriptsize

\caption{Architectural taxonomy of multi-agent LLM frameworks across graph-based, role-based, and GABM paradigms. It decomposes frameworks into core design dimensions to enable systematic, framework-level comparison independent of models.\shorten}
\vspace{-2mm}
\label{tab:framework-comparison}
\renewcommand{\arraystretch}{0.9}
\begin{tabular}{
    m{3.5cm}%
    >{\centering\arraybackslash}m{4.6cm}%
    >{\centering\arraybackslash}m{4.6cm}%
    >{\centering\arraybackslash}m{4.6cm}%
}
\toprule
\textbf{Feature} & \textbf{Graph-Based (LangGraph)} & \textbf{Role-Based (CrewAI)} & \textbf{GABM (Concordia)} \\
\midrule

\multicolumn{4}{l}{\textbf{1. Single-Agent Characteristics}} \\

\quad Main Purpose 
    & Deterministic workflow orchestration via explicit execution graphs; agent-based 
    & Task-driven coordination via role-specialized agents and delegation; task-based
    & Simulation of emergent behavior through environment-mediated agent interaction \\

\quad Agent Architecture
    & \parbox[c]{\linewidth}{\centering
        \includegraphics[width=.85\linewidth, trim=.4cm .4cm .4cm .4cm, clip]{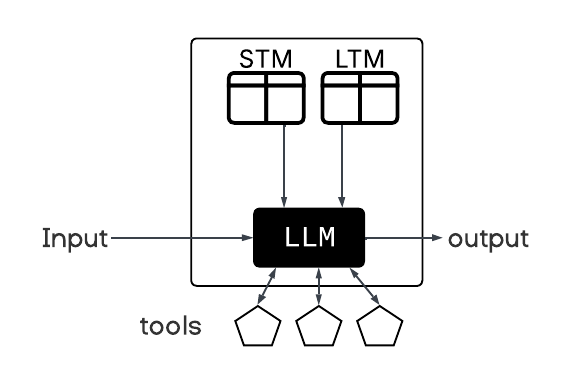}}
    & \parbox[c]{\linewidth}{\centering
        \includegraphics[width=.85\linewidth, trim=.5cm .5cm .5cm .5cm, clip]{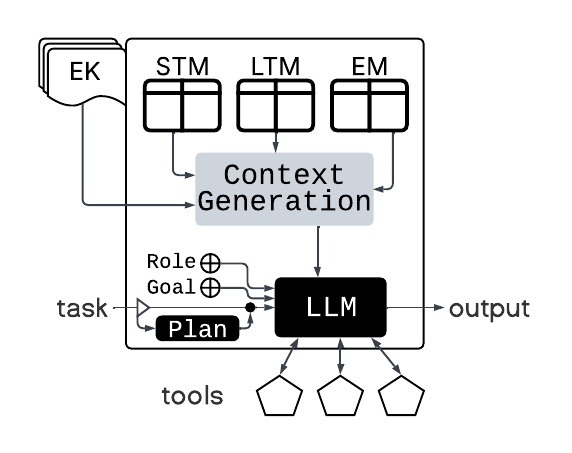}}
    & \parbox[c]{\linewidth}{\centering
        \includegraphics[width=.85\linewidth, trim=.4cm .4cm .4cm .4cm, clip]{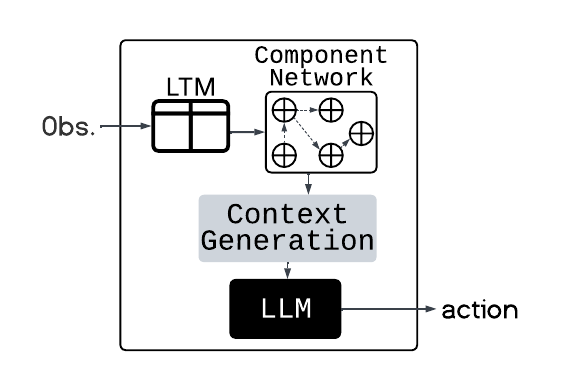}} \\

\multicolumn{4}{l}{\textbf{\quad Behavioral Specification}} \\

\quad\quad Role 
    & \xmark~Implicit; No first-class role abstraction 
    & \checkmark~Explicit; First-class role abstraction 
    & \checkmark~Optional; Provided class targets social experiments \\

\quad\quad Goal 
    & \xmark~Determined by graph; No first-class goal object 
    & \checkmark~First-class object; fixed; one per agent
    & \xmark~Environment-driven; No first-class goal object \\

\quad\quad Planning 
    & \xmark~Embedded in graph; LLM planning requires coding 
    & \checkmark~Pre-task or manager-worker planning
    & \checkmark~Optional; Provided class targets social experiments \\

\multicolumn{4}{l}{\textbf{\quad Storage}} \\

\quad\quad Long-Term Memory (LTM) 
    & \checkmark~ Retrieval-based semantic store 
    & \checkmark~Retrieval-based semantic store
    & \checkmark~LLM-queried textual memory \\

\quad\quad Short-Term Memory (STM) 
    & \checkmark Conversation or state accumulation within execution  
    & \checkmark~RAG-based working context over recent interactions 
    & \xmark~Not natively supported \\

\quad\quad Entity Memory (EM)
    & \xmark~Not natively supported 
    & \checkmark~Optional; structured entity facts via retrieval
    & \xmark~Not natively supported \\

\quad\quad Working Memory (WM)
    & \xmark~State variables must be managed by developer
    & \xmark~Implicit via task context; not explicitly modeled 
    & \checkmark~Derived from LTM; update rules by developer \\

\quad\quad External Knowledge (EK)
    & \xmark~Not natively supported 
    & \checkmark~Explicitly attachable (files, strings)   
    & \xmark~Not natively supported \\

\textbf{\quad Tool Execution}
    & \checkmark~Agent-bound tools or explicit graph tool nodes  
    & \checkmark~Agent-bound tools
    & \checkmark~Tools executed by environment, not by agents \\

\midrule

\multicolumn{4}{l}{\textbf{2. Multi-Agent Characteristics}} \\

\quad Network Topology 
    & \parbox[c]{\linewidth}{\centering
        \includegraphics[width=.7\linewidth, trim=.48cm .48cm .4cm .6cm, clip]{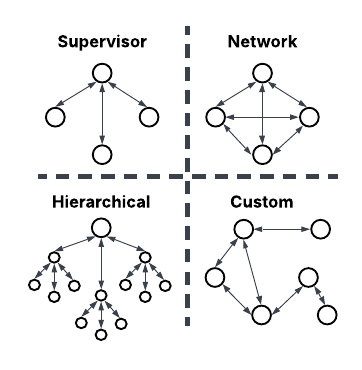}}
    & \parbox[c]{\linewidth}{\centering
        \includegraphics[width=.45\linewidth, trim=.28cm .23cm .2cm .3cm, clip]{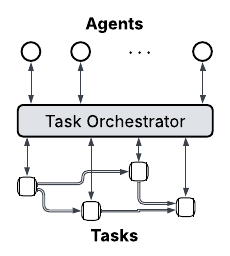}}
    & \parbox[c]{\linewidth}{\centering
        \includegraphics[width=.5\linewidth, trim=.28cm .28cm .2cm .3cm, clip]{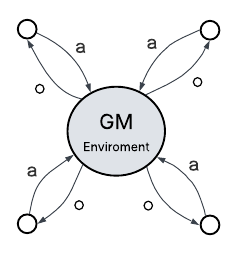}} \\

\quad  
    & \checkmark~Flexible; fixed once defined
    & \checkmark~Task hierarchy or sequence 
    & \checkmark~Fixed star topology; GM at center \\

\quad Communication Pattern 
    & \checkmark~Shared state along predefined edges 
    & \checkmark~Task-mediated; limited peer querying 
    & \xmark~Only via GM; no peer-to-peer \\

\quad Collaboration 
    & \xmark~None; procedural coordination only 
    & \checkmark~Partial; delegation-based coordination only 
    & \xmark~None; no explicit collaboration mechanisms \\

\midrule

\multicolumn{4}{l}{\textbf{3. Environment}} \\

\quad World State and Grounded Variables 
    & \xmark~None; execution context only 
    & \xmark~None; task-centric context only 
    & \checkmark~Explicit world state with grounded variables \\

\bottomrule
\end{tabular}
\vspace{-2mm}
\end{table*}

\noindent
\textbf{Definition 2.1 (Agent).} An agent is an autonomous computational entity $a = (\mathcal{R}, \mathcal{Y}, \mathcal{P}, \mathcal{S}, \mathcal{T}, f)$, where $\mathcal{R}$ is role context, $\mathcal{Y}$ is objectives, $\mathcal{P}$ is planning mechanisms, $\mathcal{S}$ is storage and knowledge resources, $\mathcal{T}$ is tools or actions, and $f$ is a reasoning function mapping observations and stored state to actions. We focus on LLM-based agents, so $f$ is an LLM. Model-native agentic capabilities, such as tool calling and internal planning, are part of $f$. Our taxonomy targets the framework-level abstractions that wrap and constrain $f$.\shorten

\noindent
\textbf{Definition 2.2 (\revthree{R3W2}{General-Purpose Multi-Agent LLM Framework}).} A general-purpose multi-agent LLM framework is an architectural system $\mathcal{F}= (\{a_{i}\}_{i=1}^{n}, \mathcal{O}, \mathcal{C}, \mathcal{E})$, where $\{a_{i}\}$ is a set of LLM-based agents, $\mathcal{O}$ is orchestration logic and control flow, $\mathcal{C}$ is communication structure and connectivity, and $\mathcal{E}$ is the shared environment whose state and transition rules mediate agent interactions.\shorten

\noindent
\textbf{Definition 2.3 (Task-Specific Multi-Agent Framework).} \revthree{R3W2}{A task-specific multi-agent framework instantiates a system-supplied orchestration loop $\mathcal{O}_{\text{fixed}}$ for one task family. It spawns sub-agents through a delegation primitive $d: a \rightarrow \{a\}$ that induces a rooted tree over $\{a_{i}\}$. Extensibility augments $\mathcal{T}$, $\mathcal{R}$, and $\mathcal{P}$ through tools, sub-agent definitions, planning modes, and hooks. Communication $\mathcal{C}$ is limited to delegation and return, and no shared environment $\mathcal{E}$ is exposed.}

\revthree{R3W2}{Coding assistants such as Claude Code~\cite{claudecode2026}, Cursor~\cite{cursor2026}, Cline~\cite{cline2026}, and OpenHands~\cite{openhands2025} instantiate Definition~2.3 (Table~\ref{tab:github-stats-updated}). Each supports sub-agent delegation but fixes $\mathcal{O}$ to a software-development loop. OpenClaw~\cite{openclaw2026} fixes $\mathcal{O}$ to a personal-assistant loop over messaging channels, so it instantiates the same definition over a different task family. This paper studies general-purpose frameworks (Definition~2.2). They expose orchestration, communication, and storage as design choices, so they admit controlled variation of one dimension at a time.}

Table~\ref{tab:framework-comparison} presents the resulting taxonomy on the representative systems of Table~\ref{tab:github-stats-updated}. It organizes design choices along two axes: architectural paradigms (Section~\ref{sec:background}) and architectural dimensions (Section~\ref{Architectural_Decisions}). Section~\ref{sec:hypotheses} then states the research questions that guide our evaluation.

\begin{table}[t]
  \centering
  \footnotesize
  \renewcommand{\arraystretch}{0.95}
  \setlength{\tabcolsep}{3.7pt}
  \caption{\revone{R1O1}{Paradigm traces on the running example (\emph{gpt-4o-mini}). Only orchestration varies. \emph{Prompts} is whether every issued prompt is retrievable. CrewAI spends a second model turn on each tool result. The skeleton matches LangGraph on calls but carries environment state in every observation.}}
  \vspace{-3mm}
  \label{tab:paradigm-traces}
  \begin{tabular}{lllrr}
    \toprule Paradigm               & Control flow  & Prompts & LLM calls & Tokens (in/out) \\
    \midrule LangGraph (graph node) & graph edges   & full    & 4         & 138\,/\,45      \\
    CrewAI (agent-bound)            & role hand-off & partial & 7         & 2{,}677\,/\,463 \\
    GABM skeleton (env)             & env-mediated  & full    & 4         & 296\,/\,204     \\
    \bottomrule
  \end{tabular}
  \vspace{-3mm}
\end{table}

\subsection{Architectural Paradigms}
\label{sec:background} Each paradigm is a coherent class of design commitments that jointly fix $\mathcal{O}$, $\mathcal{C}$, agent abstraction, and $\mathcal{E}$. \revtwo{R2O2}{Paradigms separate along a single axis: which entity holds control-flow authority, the developer at design time, the agents at runtime, or the environment through mediation. Exactly one entity holds it, so the categories are mutually exclusive by construction. Table~\ref{tab:github-stats-updated} maps the evaluated frameworks onto this axis. Task-specific frameworks fall outside it. The vendor composes authored control flow with role-conditioned delegation into a loop for one task family and ships it fixed. The developer inherits this composition, and defines sub-agents only within the slots the shipped loop provides (Definition~2.3).} \revone{R1O1}{The running example fixes one request (Boston~$\to$~New York, four days, \$500) and four roles, \emph{Planner}, \emph{Flight}, \emph{Hotel}, and \emph{Budget}, with a search tool (\emph{Web}) and a calculator (\emph{Cal}). We implement it in \emph{LangGraph}, \emph{CrewAI}, and the \emph{GABM skeleton}, holding task, tools, and model (\emph{gpt-4o-mini}) fixed. One instrumented client calls the model, so calls and tokens are comparable (Table~\ref{tab:paradigm-traces}).\footnote{Full traces are provided in the \uline{\textcolor{blue}{\href{https://github.com/CoDS-GCS/MAFBench/blob/main/supplementary_materials/APPENDICIES.pdf}{supplementary materials}}}.} Invocation is deterministic wherever the paradigm permits, so the traces isolate orchestration. Section~\ref{subsec:single_agent_chars} varies it instead.}

\subsubsection{Graph-Based Frameworks.}
Graph-based frameworks encode orchestration $\mathcal{O}$ as an explicit directed graph. Nodes represent agents or computational modules, and edges define permissible control or data flows~\cite{langgraph2025, langflow2025, n8n2025}. Execution follows this predefined structure (Figure~\ref{fig:running_example}.a). \revone{R1O1}{The developer holds control-flow authority. In our LangGraph implementation, trip state advances along author-defined edges and each agent runs once, completing the task in four LLM calls (Table~\ref{tab:paradigm-traces}). The paradigm admits two invocation modes, and Figure~\ref{fig:running_example}.a shows both: \emph{Web} bound to an agent and \emph{Cal} as a graph node. The trace places both on graph nodes, so no model call carries a tool schema and every prompt is visible. Section~\ref{subsec:single_agent_chars} varies tool invocation.}

\smallskip
\noindent
\textbf{Definition 2.4 (Graph-Based Framework).} A graph-based framework instantiates $\mathcal{O}$ as a directed graph $\mathcal{G}=(V,E)$, where $V=\{a_{1},\ldots,a_{n}\}$ are agents or components and $E\subseteq V\times V$ defines allowable control or data flows. Communication $\mathcal{C}$ propagates only along edges in $\mathcal{G}$. No explicit environment state $\mathcal{E}$ is maintained.

\subsubsection{Role-Based Frameworks.}
Role-based frameworks organize agents around textual role specifications that condition objectives, tool access, and interaction behavior~\cite{crewai2025,openaiagents2025,wu2024autogen,agno2025,OpenAgents}. Coordination emerges through role-conditioned reasoning, delegation, and structured message exchange (Figure~\ref{fig:running_example}.b). \revone{R1O1}{The agents hold control-flow authority. In our CrewAI implementation, the Planner delegates to the Flight, Hotel, and Budget roles at runtime. CrewAI binds tools to agents and offers no graph-node alternative, so each specialist spends a second model turn consuming its tool result. The task takes seven calls with an order of magnitude more input tokens (Table~\ref{tab:paradigm-traces}). Roles replace authored control flow, trading determinism for tokens. Internal delegation prompts are not retrievable, so we record task descriptions and aggregate token counts rather than every prompt.}

\smallskip
\noindent
\textbf{Definition 2.5 (Role-Based Framework).} A role-based framework instantiates agents $\{a_{i}\}_{i=1}^{n}$ with a role space $D$ and a mapping $R:\{a_{i}\}\rightarrow D$ assigning each agent a role. Role-conditioned interactions govern $\mathcal{O}$ and $\mathcal{C}$. No explicit environment state $\mathcal{E}$ is maintained.

\subsubsection{GABM Frameworks.}
GABM frameworks treat multi-agent systems as simulations where behavior emerges from \mbox{agent--environment} interactions rather than explicit orchestration. Coordination is mediated through shared world state, updated by a transition function $T$ (Figure~\ref{fig:running_example}.c). \revone{R1O2}{The environment holds control-flow authority, and the realization of $T$ is a design choice within the paradigm. Concordia~\cite{vezhnevets2023generative} realizes $T$ through the LLM. A game master interprets each agent action in narrative form and resolves its effect, adding model calls beyond those of the agents. Our skeleton realizes $T$ deterministically in code, applying a validated action without consulting the model. This bounds a round at one call per agent and completes the example in four calls with every prompt visible (Table~\ref{tab:paradigm-traces}). At equal call count it costs more input tokens than LangGraph, because each observation carries environment state where a graph edge passes only what the next node needs. Cost follows the realization, not the paradigm.}

\smallskip
\noindent
\textbf{Definition 2.6 (GABM Framework).} A GABM framework instantiates agents $\{a_{i}\}_{i=1}^{n}$ interacting with an environment state $\mathcal{E}$ governed by a transition function $T:\mathcal{E}\times\{a_{i}\}\rightarrow\mathcal{E}$. Orchestration $\mathcal{O}$ arises from environment-mediated interaction. Direct inter-agent communication $\mathcal{C}$ is not exposed.

\subsection{Architectural Design Decisions}
\label{Architectural_Decisions}

We organize architectural design decisions into three groups: single-agent characteristics (behavioral specification, storage architecture, tool execution), multi-agent characteristics (network topology, communication pattern, collaboration), and environment representation. Each group corresponds to a set of rows in Table~\ref{tab:framework-comparison}, and each row captures a distinct axis along which frameworks differ. The columns instantiate these dimensions for the three paradigms using LangGraph (graph-based), CrewAI (role-based), and Concordia (GABM) as representative systems. The agent architecture diagrams in Table~\ref{tab:framework-comparison} show how each paradigm composes storage $\mathcal{S}$, planning $\mathcal{P}$, tools $\mathcal{T}$, and the reasoning function $f$ into distinct execution pipelines. We describe each dimension below, then how they couple.

\subsubsection{Single-Agent Characteristics.}
\label{subsec:single_agent_chars}

\noindent
\textbf{Behavioral Specification.} This dimension describes how agent behavior is defined at runtime using roles $\mathcal{R}$, goals $\mathcal{Y}$, and planning $\mathcal{P}$ (Definition~2.1). \revone{R1O3}{Planning has two layers. The planning logic produces the intermediate steps and resides in $f$. The planning interface is the framework mechanism that decides whether a plan is elicited, what format it must satisfy, and where it enters the context. Frameworks wrap planning logic. They do not implement it~\cite{crewai2025, agno2025, wu2024autogen}.} The Behavioral Specification rows of Table~\ref{tab:framework-comparison} show how the three paradigms realize these components. Graph-based frameworks implement $(\mathcal{R}, \mathcal{Y}, \mathcal{P})$ procedurally within $\mathcal{O}$, so control flow determines behavior rather than explicit abstractions. \revone{R1O3}{LangGraph exposes no planning interface. The authored graph is the plan, and eliciting one from the model requires developer coding} (Figure~\ref{fig:running_example}.a)~\cite{langgraph2025,langflow2025,autogpt2025,n8n2025}. Role-based frameworks expose $\mathcal{R}$ and $\mathcal{Y}$ as first-class textual inputs that guide $f$. CrewAI adds a pre-task or manager-worker interface that requests a plan from the model and appends it to the task context (Figure~\ref{fig:running_example}.b)~\cite{crewai2025,wu2024autogen,openaiagents2025,OpenAgents}. GABM frameworks treat $(\mathcal{R}, \mathcal{Y}, \mathcal{P})$ as optional. Concordia provides role and planning classes for social simulation experiments, but the environment drives goals rather than declaring them (Figure~\ref{fig:running_example}.c)~\cite{vezhnevets2023generative}. \revone{R1O3}{These differences determine which frameworks expose a planning interface that can be varied under control.}

\smallskip
\noindent
\textbf{Storage Architecture.} This dimension captures how agent knowledge $\mathcal{S}$ is represented, accessed, and updated during execution. Let $\mathcal{S}=\{\text{LTM},\text{STM},\text{EM},\text{WM},\text{EK}\}$ denote the memory types exposed by frameworks. Each $s\in\mathcal{S}$ has four properties: lifetime (short-lived vs.\ persistent), structure (unstructured vs.\
structured), access mode (retrieval, context accumulation, or state update), and ownership (agent-local vs.\ environment-level). The Storage rows of Table~\ref{tab:framework-comparison} show which types each paradigm realizes natively. LangGraph supports retrieval-based LTM and accumulation-based STM. CrewAI adds EM and EK. Concordia centralizes storage in $\mathcal{E}$ and derives WM from LTM. Every paradigm leaves at least two types unsupported, so the available access mechanisms are retrieval, accumulation, and their hybrid. Memory types requiring custom implementation are excluded.\shorten

\smallskip
\noindent
\textbf{Tool Execution Model.} This dimension determines how external actions in $\mathcal{T}$ are invoked. The Tool Execution row of Table~\ref{tab:framework-comparison} shows three realizations: agent-bound tools or graph tool nodes constrained by $\mathcal{O}$ in graph-based frameworks (\emph{Web} and \emph{Cal} in Figure~\ref{fig:running_example}.a), agent-bound tools through role assignment in role-based ones (Figure~\ref{fig:running_example}.b), and execution routed through $\mathcal{E}$ in GABM frameworks, where agent outputs become state updates (Figure~\ref{fig:running_example}.c).

\revthree{R3Q3}{Every paradigm delegates tool selection to $f$, so the architectural property is not who selects but whether invocation is guaranteed. A graph tool node and a mediator applying an action invoke deterministically. Authoring the node and its position in $\mathcal{O}$ costs developer effort, but guarantees execution and saves the tokens spent transmitting schemas and deciding invocation. Agent-bound invocation carries no such guarantee. Two instructions for the same LangGraph configuration give opposite outcomes under \emph{gpt-4o-mini}. A permissive one offering the tool as optional never calls it and yields no valid plan. A directive one naming it as the way to obtain prices invokes it and completes. CrewAI uses the directive form and re-enters the model after each tool result, biasing its loop toward invoking, and \emph{Opus~5} invokes under both.\footnote{Per-configuration invocation results are in the \uline{\textcolor{blue}{\href{https://github.com/CoDS-GCS/MAFBench/blob/main/supplementary_materials/APPENDICIES.pdf}{supplementary materials}}}.} Invocation depends on wording, agent loop, and model at once, so measuring agent-bound selection would rank the model rather than isolate a design dimension.}\revone{R1O1}{}

\subsubsection{Multi-Agent Characteristics}
\label{sec:multi_agent}

The bottom rows of Table~\ref{tab:framework-comparison} define multi-agent interaction through three aspects: network topology (the structure of $\mathcal{C}\subseteq \{a_{i}\}\times\{a_{i}\}$), communication pattern (how information is exchanged over $\mathcal{C}$), and collaboration (how agents work toward shared goals). The paradigms differ on all three. Graph-based frameworks define $\mathcal{C}$ explicitly at design time and fix it during execution, so communication is shared state along predefined edges (Figure~\ref{fig:running_example}.a)~\cite{langgraph2025,autogpt2025,n8n2025}. Role-based frameworks induce $\mathcal{C}$ from roles $\mathcal{R}$ and objectives $\mathcal{Y}$, so communication is task-mediated and collaboration is delegation-based (Figure~\ref{fig:running_example}.b)~\cite{crewai2025,wu2024autogen,openaiagents2025}. GABM frameworks remove direct agent connections ($\mathcal{C}=\emptyset$) and route everything through a Game Master at the center of a fixed star topology (Figure~\ref{fig:running_example}.c)~\cite{vezhnevets2023generative}. Neither graph-based nor GABM frameworks provide explicit collaboration mechanisms. These structural constraints determine how information flows between agents and whether coordination scales with the number of agents.\shorten

\subsubsection{Environment.}
\label{sec:environment} This dimension describes how execution is mediated through shared world state $\mathcal{E}$ (Definition~2.2). The Environment row of Table~\ref{tab:framework-comparison} shows that only GABM frameworks expose one. Graph-based frameworks operate on execution context, where intermediate outputs propagate along workflow edges. Role-based frameworks maintain task-centric context without persistent global state. GABM frameworks define an explicit $\mathcal{E}$ with grounded variables that persist across rounds, updated by a transition function $T:\mathcal{E}\times\{a_{i}\}\rightarrow\mathcal{E}$ (Definition~2.6)~\cite{vezhnevets2023generative}. The presence or absence of $\mathcal{E}$ determines whether coordination state survives across rounds. Frameworks without $\mathcal{E}$ must re-derive shared context from message history, but GABM frameworks accumulate it in persistent world state.\shorten

\subsubsection{Dimension Coupling and Evaluation Scope}
\label{sec:dim_coupling} \revtwo{R2O2}{Choosing a paradigm fixes some dimensions and leaves others open, so they are not independent. A dimension is \emph{fixed} when the paradigm or framework determines it, and \emph{variable} when the developer selects it, whether freely or from natively supported options. Orchestration $\mathcal{O}$ and communication $\mathcal{C}$ are coupled throughout. The execution graph is also the communication structure, the role mapping $\mathcal{R}$ induces both and binds $\mathcal{T}$ to agents, and $\mathcal{E}$ mediates $\mathcal{C}$ and owns $\mathcal{S}$.} \revthree{R3Q1}{Controlled variation needs one variable dimension while the rest stay fixed. We therefore vary topology in graph-based frameworks, storage in graph-based and role-based ones, planning interface in CrewAI, and tools in LangGraph. Varying a definitional dimension would change the paradigm rather than isolate it, so some framework--dimension combinations are absent from the evaluation grid. Task-specific frameworks seal $\mathcal{O}$, $\mathcal{C}$, and $\mathcal{S}$ inside the system loop, leaving only $\mathcal{P}$, $\mathcal{R}$, and $\mathcal{T}$. Their planning offers the same interfaces we already evaluate. A full factorial study and a dedicated study of task-specific systems follow directly from this scoping.\shorten }

\subsection{\revthree{R3M1}{Research Questions}}
\label{sec:hypotheses} \revthree{R3M1}{The taxonomy raises one open question per dimension, each derived from the design differences above and answered in}~Section~\ref{sec:eval}.

\smallskip
\noindent
\textbf{RQ1 (Memory Access Mode).} \revthree{R3M1}{Which access mode governs multi-hop retrieval (RQ1a), long-range understanding (RQ1b), and in-session learning (RQ1c)? Does hybrid retrieval--accumulation behave like either component alone? And what does each architecture support under knowledge revision (RQ1d), a state update rather than an append?}

\smallskip
\noindent
\textbf{RQ2 (Planning Interface).} \revthree{R3M1}{How does the output format a planning interface imposes on $\mathcal{P}$ affect accuracy and runtime, and does the effect come from reasoning or parsing failures?}\shorten

\smallskip
\noindent
\textbf{RQ3 (Specialization).} \revthree{R3M1}{Does a role activate domain reasoning by naming an identity or by binding a capability, and does binding require procedure?}

\smallskip
\noindent
\textbf{RQ4 (Communication Topology).} \revthree{R3M1}{How does connectivity in $\mathcal{C}$ affect coordination as agents grow in number, and what connectivity do different tasks need within bounded rounds?}

\smallskip
\noindent
\textbf{RQ5 (Orchestration Overhead).} \revthree{R3M1}{What execution cost does each orchestration structure impose on a trivial task, and how much of the cost of mediation through $\mathcal{E}$ is inherent to the paradigm rather than a framework's implementation?}
  \section{\NoCaseChange{MAFBench}: A Unified Benchmark}
\label{sec:benchmark_collection}

This section introduces \textbf{MAFBench}, a unified benchmark for evaluating
multi-agent frameworks across the architectural dimensions in
Section~\ref{sec:taxonomy_analysis}.\shorten

\smallskip
\noindent\textbf{MAFBench Contributions.}
Existing benchmarks evaluate individual agent capabilities in isolation, not
framework-level design choices. \revtwo{R2O3}{MAFBench compares frameworks as
deployed, standardizing everything around the framework and leaving the
framework itself untouched.} It contributes four pieces of infrastructure. A
\emph{unified execution pipeline} wraps heterogeneous frameworks behind a common
interface, enforcing identical prompts and stopping criteria. \emph{LLM-based
semantic scoring} replaces brittle string matching and handles diverse answer
formats. \emph{Configuration parity} centralizes model selection, context limits,
batching, and concurrency, so observed differences reflect architecture rather
than drift. \revthree{R3W3}{Every framework exposing decoding gets the same
setting within an experiment, at or near temperature $0$. Frameworks exposing
none use provider defaults, and no seed is set.} \emph{Cost control and backend
routing} redirect compatible API calls to alternative providers without
modifying framework code. \revtwo{R2O3}{Where a paradigm has one production
framework, its measurements carry that implementation's idiosyncrasies, so we
add a minimal reference implementation to separate paradigm cost from
implementation cost (Section~\ref{subsec:overhead}).} Each subsection below adds
only what its dimension requires.

\subsection{Memory Benchmarks}
\label{subsec:bench_memory}

Prior memory benchmarks for LLM-based agents evaluate isolated capabilities:
long-context retrieval~\cite{hsieh2024ruler,wu2024longmemeval,zhang2024bench},
dialogue and state tracking~\cite{li2018towards,liu2021benchmarking},
incremental preference
learning~\cite{casanueva2020efficient,larson2019evaluation,he2023large}, and
controlled knowledge revision~\cite{zhong2023mquake}. None captures how agents
retain, integrate, generalize, and revise information across multi-turn
interactions. We therefore adopt
\emph{MemoryAgentBench}~\cite{hu2025memoryagentbench}, which unifies these
dimensions into four core competencies.\footnote{\url{https://huggingface.co/datasets/ai-hyz/MemoryAgentBench}}

\smallskip
\noindent\textbf{Benchmark Unified Interface.}
MemoryAgentBench organizes evaluation into four competency-aligned splits
composed of adapted subtasks (Table~\ref{tab:memory-bench}). \textit{Accurate
Retrieval (AR)} measures factual recall and multi-hop reasoning over long
contexts. \textit{Test-Time Learning (TTL)} evaluates in-session learning of
concepts and preferences. \textit{Long-Range Understanding (LRU)} targets
long-range abstraction and cross-document reasoning. \textit{Selective
Forgetting (SF)} probes controlled memory revision through counterfactual
updates~\cite{hu2025memoryagentbench}.

\smallskip
\noindent\textbf{Evaluation Setup.}
We standardize document ingestion using fixed-size text chunking with controlled
overlap, consistent ordering, and identical session initialization across
frameworks. The suite is long-context, so we route memory-intensive workloads to
Groq-hosted models~\cite{groq} through transparent backend routing. This lowers
evaluation cost without modifying framework implementations.

\smallskip
\noindent\textbf{Scope: framework-native memory.}
\revthree{R3W4}{We evaluate the memory architectures the frameworks provide
themselves. Plugin layers such as Mem0~\cite{chhikara2025mem0},
Letta~\cite{letta}, Zep~\cite{rasmussen2025zep}, MIRIX~\cite{wang2025mirix}, and
EverMemOS~\cite{evermemos} attach to any framework, so they do not differentiate
frameworks along the dimension we study. 
MAFBench's adapter interface holds the framework fixed while varying the memory backend, so it enables their in-depth evaluation.
}

\begin{table}[t]\centering
\caption{MemoryAgentBench statistics per competency split. Each split evaluates a distinct memory capability: Accurate Retrieval (AR), Test-Time Learning (TTL), Long-Range Understanding (LRU), and Selective Forgetting (SF). Question counts and context lengths are in thousands of words.}
\vspace{-3mm}
\label{tab:memory-bench}
\footnotesize
\begin{tabular}{l r r | rrrr | rrrr}
\toprule
 & & & \multicolumn{4}{c|}{\textbf{Questions}} & \multicolumn{4}{c}{\textbf{Context (k words)}} \\
Split & Sessions & Subtasks & Min & Max & Avg & Total & Min & Max & Avg & Total \\
\midrule
AR & 22 & 4 & 60 & 100 & 90.9 & 2,000 & 50 & 560 & 206 & 4,551 \\
TTL & 6 & 2 & 100 & 200 & 116.7 & 700 & 69 & 1,040 & 236 & 1,417 \\
LRU & 110 & 2 & 1 & 10 & 1.6 & 171 & 48 & 560 & 129 & 14,281 \\
SF & 8 & 2 & 100 & 100 & 100.0 & 800 & 4 & 183 & 63 & 511 \\
\bottomrule
\end{tabular}
\vspace{-2mm}
\end{table} 
\begin{table}[t]
\centering
\footnotesize
\caption{Planning benchmark statistics. GSM8K covers numerical reasoning, CSQA covers commonsense reasoning, and MATH covers symbolic reasoning.}
\vspace{-3mm}
\label{tab:benchmark_stats_multicol}
\begin{tabular}{l r | rrrr | rrrr}
\toprule
& & \multicolumn{4}{c|}{\textbf{Question Length (words)}} &
    \multicolumn{4}{c}{\textbf{Gold Answer Length (words)}} \\
\cmidrule(lr){3-6} \cmidrule(lr){7-10}
\textbf{Benchmark} & \textbf{\#Q} &
\textbf{Min} & \textbf{Max} & \textbf{Avg} & \textbf{Total} &
\textbf{Min} & \textbf{Max} & \textbf{Avg} & \textbf{Total} \\
\midrule
GSM8K~\cite{cobbe2021training} & 1319 & 15 & 164 & 46.3 & 61,005 & 1 & 1 & 1.0 & 1,319 \\
CSQA~\cite{talmor-etal-2019-commonsenseqa} & 1221 & 19 & 69 & 31.6 & 38,615 & 1 & 1 & 1.0 & 1,221 \\
MATH~\cite{hendrycks2measuring}  & 5000 & 2 & 252 & 30.7 & 153,261 & 1 & 19 & 1.3 & 6,345 \\
\bottomrule
\end{tabular}
\vspace{-2mm}
\end{table} 

\subsection{Planning Interface Benchmarks}
\label{subsec:bench_planning}
This study varies the planning interface (Section~\ref{subsec:single_agent_chars})
and holds the planning logic fixed. \revone{R1O3}{The same model generates the
plan under every mode. CrewAI~\cite{crewai2025} is the only evaluated framework
exposing a runtime planning interface, so we compare it against no interface and
against a free-form interface injected outside the framework.} Evaluation of
these interfaces remains fragmented across reasoning benchmarks. We adopt
\emph{GSM8K}~\cite{cobbe2021training},
\emph{CommonsenseQA (CSQA)}~\cite{talmor-etal-2019-commonsenseqa}, and
\emph{MATH}~\cite{hendrycks2measuring} (Table~\ref{tab:benchmark_stats_multicol}),
which cover numerical, commonsense, and symbolic reasoning tasks where
multi-step planning influences
outcomes.\footnote{\url{https://huggingface.co/datasets/openai/gsm8k} (test only);
\url{https://huggingface.co/datasets/tau/commonsense_qa} (validation only);
\url{https://github.com/hendrycks/math}}

\noindent\textbf{Benchmark Planning Interfaces.}
Each benchmark runs under three controlled interfaces. \emph{NoPlan} passes tasks
directly to the LLM without eliciting a plan. \emph{Crew-Plan} follows CrewAI's
schema-constrained two-stage interface~\cite{crewai2025}, where the framework
requests a structured plan from the LLM and appends it to the task description.
\emph{Direct-LLM-Plan} elicits an unconstrained natural-language plan from the
same model and injects it into the context before answer generation.
\revone{R1O3}{The three modes differ in whether a plan is requested, what format
it must satisfy, and where it enters the context. They do not differ in which
model produces the plan.}

\noindent\textbf{Evaluation Setup.}
The pipeline treats the planning interface as a configurable execution stage and
supports complexity-preserving subsampling. We use a 100-problem MATH subset.
Formatting failures are tracked separately from reasoning errors, and all three
interfaces run across multiple LLM backends to verify that observed effects are
architectural rather than model-specific. No interface verifies, critiques, or
replans. Each plan is produced once and injected once, so differences come from
format and injection point rather than control policy. CrewAI builds its plan
internally and exposes no callback, so we report its schema-violation and
runtime effects. Free-form plans are logged in
full.\footnote{Annotated plans for the same question under each interface appear
in the supplement.}

\subsection{Specialization Benchmarks}
\label{subsec:bench_specialization}
Specialization in LLM-based agents refers to activating domain-relevant
reasoning through conditioning. \revone{R1O4}{We examine whether a role names an
identity or binds a capability, and which structures the domain-specific
reasoning encoded in the model.} Using machine learning tasks from
CatDB~\cite{catdb}, we measure behavioral differences with fixed data and models.

\noindent\textbf{Benchmark Structure and Conditioning Variants.}
The benchmark comprises five public datasets spanning regression, binary, and
multiclass classification (Table~\ref{tab:datasets}), \revone{R1O4}{each evaluated under four
conditioning strategies.} \emph{Role-based prompting} assigns a professional
identity in the task description. \emph{Planning-based conditioning} adds an
intermediate step where the LLM generates a solution plan before code
generation. \emph{Expert-guided conditioning} injects methodological instructions
reflecting data-science workflows. \revone{R1O4}{\emph{Dataset-profile tooling}
grants the agent a tool returning feature types, cardinality, and missingness,
without prescribing a workflow. These variants isolate identity framing,
reasoning structure, procedural guidance, and data access.}\shorten

\noindent\textbf{Evaluation Setup.}
Task structure, data loading, model execution, and metrics stay fixed. Only the
conditioning strategy varies across runs. Each agent generates a data-centric ML
pipeline, and we score the executed pipeline rather than the generated code. We
report MAE for regression and precision, recall, and F1 for classification, so
our results compare conditioning strategies on task
performance rather than on output style.

\setlength{\tabcolsep}{7pt}
\begin{table}[t]
\centering
\renewcommand{\arraystretch}{.85}
\footnotesize
\caption{Specialization benchmark datasets from CatDB~\cite{catdb}, spanning regression, binary, and multiclass classification tasks with varying feature dimensionality.}
\vspace{-3mm}
\label{tab:datasets}
\begin{tabular}{lccc}
\hline
\textbf{Dataset} & \textbf{Task Type} & \textbf{Target Variable} & \textbf{\# Features} \\
\hline
Utility   & Regression                & CSRI        & 12  \\
WiFi      & Binary Classification      & TechCenter  & 9   \\
EU-IT     & Multiclass Classification  & Position    & 23  \\
Yelp      & Multiclass Classification  & stars       & 184   \\
Volkert   & Multiclass Classification  & class       & 181   \\
\hline
\end{tabular}
\vspace{-5mm}
\end{table} 

\subsection{Tool Use Benchmarks and Limitations}
\label{sec:tools-benchmark}

Tool use enables LLM-based agents to interact with external APIs and services
beyond language generation. Recent benchmarks such as ToolBench and its
deterministic variant
StableToolBench~\cite{guo2024stabletoolbench,qin2025tool,qintoolllm} evaluate an
agent's ability to select and invoke appropriate tools from large, heterogeneous
registries.

\noindent\textbf{Benchmark Structure and Scope.}
Figure~\ref{fig:stabletoolbench-overview} shows how StableToolBench spans
multiple domains and interface complexities, including single-endpoint tools and
multi-endpoint APIs~\cite{guo2024stabletoolbench}. It stresses correct tool
selection, parameter grounding, and multi-step interaction with closely related
endpoints. MAFBench integrates it through a tool-use evaluation layer that
preserves the original queries, server, and scoring logic. A pre-execution
selection step bounds the tool subset per query before framework binding, so
tool sets are identical across runs. \revthree{R3Q3}{We do not report
quantitative tool-use results, because current frameworks delegate tool
invocation entirely to the LLM without architectural control over selection.
Tool-use evaluation is deferred until frameworks introduce explicit
orchestration mechanisms.}\shorten

\subsection{Coordination and Scaling Benchmarks}
\label{subsec:bench_coordination}

Existing frameworks support agent execution and task decomposition, but they
lack native multi-round peer-to-peer communication over arbitrary topologies.
Interaction topology therefore dictates coordination rather than explicit
framework mechanisms. We adopt
AGENTSNET~\cite{coordinationcollaborativereasoning}, which analyzes emergent
collective behavior under constrained message
passing.\footnote{Dataset: \url{https://huggingface.co/datasets/disco-eth/AgentsNet}; \\
Code: \url{https://github.com/floriangroetschla/AgentsNet}} It evaluates
\emph{local consistency} and \emph{global alignment} under limited connectivity.

\begin{figure}[t]
\vspace{-1mm}
\centering
\begin{tikzpicture}

\pgfplotsset{
    common style/.style={
        title style={font=\tiny\bfseries, yshift=-8pt},
        label style={font=\tiny},
        tick label style={font=\tiny},
        ymajorgrids=true,
        grid style={dashed, gray!30},
        axis line style={gray!60},
        ylabel style={font=\tiny, xshift=5pt}, 
        xlabel style={font=\tiny, yshift=2pt},
        every axis plot/.append style={
            draw=black!70,
            line width=0.4pt
        }
    }
}

\hspace{-4mm}
\begin{axis}[
    common style,
    name=qlen,
    ybar,
    bar width=4pt,
    width=0.39\columnwidth,
    height=3.cm,
    title={(a)},
    xlabel={Words},
    ylabel={Count},
    ylabel style={xshift=-6pt, yshift=-5pt},
    xtick={1,2,3,4,5},
    xticklabels={0-20,20-40,40-60,60-80,80+},
    xticklabel style={rotate=90, anchor=east},
    ymin=0, ymax=650,
    enlarge x limits=0.2,
    nodes near coords,
    nodes near coords style={font=\tiny, gray!80, /pgf/number format/fixed, yshift=-2pt}
]

\addplot [pattern=north east lines, pattern color=blue!40!black!60] coordinates {
    (1,6) (2,246) (3,458) (4,52) (5,3)
};
\end{axis}

\hspace{2mm}
\begin{axis}[
    common style,
    name=apitools,
    at={(qlen.east)}, anchor=west, xshift=6mm, 
    ybar,
    bar width=4pt,
    width=0.39\columnwidth,
    height=3.cm,
    title={(b)},
    xlabel={Endpoints},
    ylabel={\# Tools},
    ylabel style={xshift=-6pt, yshift=-5pt},
    xtick={1,2,3,4,5,6},
    xticklabels={1,2-5,6-10,11-20,21-50,50+},
    xticklabel style={rotate=90, anchor=east},
    ymin=0, ymax=650,
    enlarge x limits=0.15,
    nodes near coords,
    nodes near coords style={font=\tiny, gray!80}
]
\addplot [pattern=crosshatch, pattern color=red!40!black!60] coordinates {
    (1,436) (2,182) (3,18) (4,13) (5,8) (6,1)
};
\end{axis}
\hspace{5mm}

\begin{axis}[
    common style,
    name=cats,
    at={(apitools.east)}, anchor=west, xshift=10mm,
    xbar,
    bar width=3.pt,
    width=0.44\columnwidth,
    height=3.2cm,
    title={(c)},
    xlabel={Query Count},
    symbolic y coords={
        Business,Media,Social,Sports,Entertainment,
        Video/Images,Data,Finance,Movies,Tools
    },
    ytick=data,
    yticklabel style={font=\tiny},
    ylabel style={xshift=0pt}, 
    xmin=0, xmax=950,
    xmajorgrids=true,
    ymajorgrids=false,
    nodes near coords,
    nodes near coords style={font=\tiny, xshift=1pt, anchor=west}
]
\addplot [pattern=dots, pattern color=green!40!black!60] coordinates {
    (96,Business) (111,Media) (152,Social) (183,Sports)
    (213,Entertainment) (244,Video/Images) (272,Data)
    (275,Finance) (413,Movies) (746,Tools)
};
\end{axis}

\end{tikzpicture}
\vspace{-4mm}
\caption{Characteristics of StableToolBench, showing the distributions of (a) query length, (b) number of APIs per tool, and (c) query counts across tool categories.}
\label{fig:stabletoolbench-overview}
\vspace{-1mm}
\end{figure}

\noindent\textbf{Benchmark Structure and Conditioning Variants.}
AGENTSNET includes five coordination tasks: \emph{Coloring} (local conflict
resolution), \emph{Matching} (bilateral agreement), \emph{VertexCover} (global
resource minimization), \emph{LeaderElection} (symmetry breaking), and
\emph{Consensus} (long-range information
propagation)~\cite{coordinationcollaborativereasoning}. Agents interact only
along edges of predefined graph families:
\emph{Small-World}~\cite{watts1998collective},
\emph{Scale-Free}~\cite{scale_free}, and \emph{Delaunay}~\cite{delaunay}. This
isolates how clustering, path length, and connectivity heterogeneity affect
coordination dynamics.

\noindent\textbf{Evaluation Setup.}
To reflect the interaction constraints of role-based and GABM frameworks, we
implement a topology rewriting engine that derives sequential, hierarchical, and
centralized structures from shared base graphs while preserving agent sets. The
centralized structure mediates interaction through a communication hub that
stands in for the GABM environment. We then execute identical coordination tasks
under each pattern using consistent prompts, models, and task semantics, and
sweep across tasks, graph families, and agent counts.
  \section{Evaluation}
\label{sec:eval}

\begin{table}[t]
\vspace{-2mm}
\centering
\scriptsize
\renewcommand{\arraystretch}{0.95}
\setlength{\tabcolsep}{1pt}
\caption{MAFBench evaluation summary. Each row maps an architectural dimension
to its benchmark, metrics, and \revthree{R3M1}{research question}
(Section~\ref{sec:hypotheses}). One dimension varies per experiment while all
others stay fixed.}
\vspace{-4mm}
\label{tab:eval_summary}
\begin{tabular}{@{}p{2.3cm} p{2.3cm} p{2.9cm} l@{}}
\toprule
\textbf{Dimension} & \textbf{Benchmark} & \textbf{Metrics} & \textbf{RQ} \\
\midrule
Memory architecture & MemoryAgentBench~\cite{hu2025memoryagentbench} & Accuracy, F1, Recall@5 & \revthree{R3M1}{RQ1} \\
Planning interface & GSM8K~\cite{cobbe2021training}, CSQA~\cite{talmor-etal-2019-commonsenseqa}, MATH~\cite{hendrycks2measuring} & Accuracy, failure \mbox{rate}, \mbox{runtime} & \revthree{R3M1}{RQ2} \\
Specialization & CatDB tasks~\cite{catdb} & Precision, Recall, F1 & \revthree{R3M1}{RQ3} \\
Communication topology & \textsc{AgentsNet}~\cite{coordinationcollaborativereasoning} & Success, rounds, tokens, \mbox{runtime} & \revthree{R3M1}{RQ4} \\
Orchestration \mbox{overhead} & Trivial query (50 trials) & Latency, throughput, \mbox{output} size & \revthree{R3M1}{RQ5} \\
\bottomrule
\end{tabular}
\vspace{-3mm}
\end{table}

Each study varies a single framework-level dimension using our MAFBench
pipeline, holding models, prompts, data, and execution semantics fixed.
\revtwo{R2O1}{Frameworks provide mechanisms at two levels. Per-agent mechanisms
configure one agent: storage, planning interface, specialization, and the
orchestration wrapper around a model call. Cross-agent mechanisms govern
interaction: communication topology and coordination. We isolate the per-agent
level first (Section~\ref{subsec:single_agent}), so cross-agent
differences (Section~\ref{subsec:multi_agent}) follow from coordination rather
than confounded per-agent machinery.} Table~\ref{tab:eval_summary} maps each
dimension to its benchmark and metrics. MemoryAgentBench aggregates
heterogeneous subtask metrics as category-level
means~\cite{hu2025memoryagentbench}. \revthree{R3W3}{To manage cost, each
configuration runs once. The planning interface is repeated five times on four
models to bound variance (Section~\ref{subsec:threats}).}

\vspace{-2mm}
\subsection{\revtwo{R2O1}{Single-Agent and Overhead Evaluation}}
\label{subsec:single_agent}
We first examine how framework orchestration, memory, and planning affect
execution cost and reasoning behavior.

\subsubsection{Framework Overhead}
\label{subsec:overhead}
We evaluate how orchestration architecture affects baseline execution cost,
independent of memory, planning, tool use, and coordination. We fix the LLM
(\emph{gpt-4o-mini}), the query (\emph{``What is 2+2?''}), prompts, and
concurrency, and vary only the orchestration layer under minimal viable
configurations: a raw \emph{Direct LLM} baseline; graph- and role-based
\emph{LangGraph}~\cite{langgraph2025}, \emph{OpenAgents}~\cite{OpenAgents},
\emph{AutoGen}~\cite{wu2024autogen}\revthree{R3Q2}{\footnote{AutoGen forked into
AG2 (\texttt{pyautogen}) and an event-driven rewrite
(\texttt{autogen-agentchat}). We evaluate \texttt{autogen-agentchat} 0.7.5
throughout. Per-experiment versions are pinned in the artifact.}},
\emph{CrewAI}~\cite{crewai2025}, \emph{Agno}~\cite{agno2025}, and the
\emph{OpenAI SDK}~\cite{openaiagents2025}; the GABM framework
\emph{Concordia}~\cite{vezhnevets2023generative}; \revone{R1O2}{a \emph{minimal
GABM skeleton} we implement to separate paradigm cost from implementation cost}\revtwo{R2O4}{};
and \revthree{R3W2}{\emph{OpenHands}~\cite{openhands2025} as an out-of-taxonomy
reference}. We run 50 trials and report median (p50) and tail (p95) latency,
throughput, and output size.

\captionsetup[subfigure]{font=scriptsize}
\begin{figure}[t]
  \begin{subfigure}
    {0.57\linewidth}
    \begin{tikzpicture}
      \begin{groupplot}
        [ group style={group size=1 by 2, vertical sep=0.4cm, x descriptions at=edge bottom},
        width=1.25\linewidth, symbolic x coords={LLM,Sk,OA,LG,OH,CA,Ag,AG,OS,Co},
        xtick=data, ybar, /pgf/bar width=2.6pt, enlarge x limits=0.10, ymajorgrids=true,
        grid style={dashed,gray!30}, tick label style={font=\tiny}, ylabel style={font=\tiny},
        xlabel style={font=\tiny}, every axis plot/.append style={fill opacity=0.9}
        ] \nextgroupplot[height=2.5cm, ymin=28, ymax=40, ytick={30,35}, axis x
        line*=top, legend style={at={(0.02,0.98)}, anchor=north west, font=\tiny, legend columns=-1},
        ylabel={Latency (s)}, ylabel style={at={(axis description cs:1.3,-1.2)}, anchor=west}]
        \addplot+[fill=blue!60] coordinates { (LLM,0.495)(Sk,0.522)(OA,0.527)(LG,0.548)(OH,0.657)(CA,0.741)(Ag,0.902)(AG,0.964)(OS,1.275)(Co,30.564)};
        \addplot+[fill=green!60] coordinates
        { (LLM,0.774)(Sk,0.833)(OA,0.751)(LG,0.789)(OH,0.951)(CA,0.936)(Ag,1.134)(AG,1.255)(OS,1.783)(Co,37.183)};
        \legend{p50, p95} \nextgroupplot[height=2.3cm, ymin=0, ymax=2, ytick={0,1,2},
        axis x line*=bottom, xticklabel style={rotate=90, anchor=east, font=\tiny}]
        \addplot+[fill=blue!60] coordinates
        { (LLM,0.495)(Sk,0.522)(OA,0.527)(LG,0.548)(OH,0.657)(CA,0.741)(Ag,0.902)(AG,0.964)(OS,1.275)(Co,30.564)};
        \addplot+[fill=green!60] coordinates { (LLM,0.774)(Sk,0.833)(OA,0.751)(LG,0.789)(OH,0.951)(CA,0.936)(Ag,1.134)(AG,1.255)(OS,1.783)(Co,37.183)};
      \end{groupplot}
    \end{tikzpicture}
    \caption{Latency}
  \end{subfigure}
  \begin{subfigure}
    {0.42\linewidth}
    \begin{tikzpicture}
      \begin{axis}[
        width=1.15\linewidth,
        height=3.6cm,
        ymin=0,
        ymax=8,
        symbolic x coords={LLM,Sk,OA,LG,OH,CA,Ag,AG,OS,Co},
        xtick=data,
        xticklabel style={rotate=90, anchor=east, font=\tiny},
        ylabel={Throughput (req/s)},
        ylabel style={at={(axis description cs:1.55,0.)}, anchor=west, font=\tiny},
        tick label style={font=\tiny},
        xlabel style={font=\tiny},
        ybar,
        bar width=2.6pt,
        enlarge x limits=0.10,
        ymajorgrids=true,
        grid style={dashed,gray!30},
        every axis plot/.append style={fill opacity=0.9}
      ]
        \addplot+[fill=red!60] coordinates { (LLM,7.046)(Sk,5.391)(OA,5.831)(LG,2.896)(OH,5.469)(CA,5.040)(Ag,4.167)(AG,3.707)(OS,2.680)(Co,0.124)};
      \end{axis}
    \end{tikzpicture}
    \caption{Throughput}
  \end{subfigure}

  \hspace{-2mm}
  \begin{subfigure}
    {0.42\linewidth}
    \begin{tikzpicture}
      \begin{semilogyaxis}
        [ width=1.15\linewidth, height=3.5cm, ymode=log, log basis y=10, ymin=1,
        ymax=1e7, symbolic x coords={LLM,Sk,OA,LG,OH,CA,Ag,AG,OS,Co}, xtick=data,
        xticklabel style={rotate=90, anchor=east, font=\tiny}, ylabel={Output size (chars)},
        ylabel style={at={(axis description cs:1.55,0.)}, anchor=west, font=\tiny},
        tick label style={font=\tiny}, xlabel style={font=\tiny}, ybar, bar width=2.6pt,
        enlarge x limits=0.10, ymajorgrids=true, grid style={dashed,gray!30},
        every axis plot/.append style={fill opacity=0.9}] \addplot+[fill=orange!60]
        coordinates
        { (LLM,1.0)(Sk,1.0)(OA,14.9)(LG,14.92)(OH,15.0)(CA,28.54)(Ag,14.9)(AG,15.0)(OS,14.1)(Co,100567.0)};
      \end{semilogyaxis}
    \end{tikzpicture}
    \caption{Mean output}
  \end{subfigure}
  \hspace{-4mm}
  \begin{subfigure}
    {0.57\linewidth}
    \begin{tikzpicture}
      \begin{semilogyaxis}
        [ width=1.15\linewidth, height=3.5cm, ymode=log, log basis y=10, ymin=10,
        ymax=8000, symbolic x coords={LLM,Sk,OA,LG,OH,CA,Ag,AG,OS,Co}, xtick=data,
        xticklabel style={rotate=90, anchor=east, font=\tiny}, ylabel={Input tokens},
        ylabel style={at={(axis description cs:1.35,0.)}, anchor=west, font=\tiny},
        tick label style={font=\tiny}, xlabel style={font=\tiny}, ybar, bar width=2.6pt,
        enlarge x limits=0.10, ymajorgrids=true, grid style={dashed,gray!30},
        every axis plot/.append style={fill opacity=0.9}] \addplot+[fill=violet!60]
        coordinates
        { (LLM,30)(Sk,46)(OA,14)(LG,14)(OH,2914)(CA,168)(Ag,14)(AG,26)(OS,25)(Co,3150)};
      \end{semilogyaxis}
    \end{tikzpicture}
    \caption{Input tokens}
  \end{subfigure}
  \hfill
  \noindent
  \fbox{\parbox{0.85\linewidth}{\scriptsize \textbf{Abbreviations:} LLM\,=\,Direct
  LLM; Sk\,=\,GABM skeleton; OA\,=\,OpenAgents; LG\,=\,LangGraph; OH\,=\,OpenHands$^{\dagger}$;
  CA\,=\,CrewAI; Ag\,=\,Agno; AG\,=\,AutoGen; OS\,=\,OpenAI SDK; Co\,=\,Concordia.
  \quad $^{\dagger}$out-of-taxonomy reference.}}
  \caption{Orchestration overhead on a trivial task (``What is 2+2?'') over 50 trials,
  ordered by p50 latency (Sk completed 46, OS 49). Graph- and role-based frameworks
  add modest overhead. \revone{R1O2}{The minimal GABM skeleton does too, so Concordia's orders-of-magnitude higher latency, lower throughput, and larger output are implementation, not paradigm, cost.}}
  \label{fig:framework-overhead}
\end{figure}

Figure~\ref{fig:framework-overhead} compares each to the Direct LLM baseline
(p50 $495$\,ms, $7.0$~req/s), which has the lowest latency and highest
throughput. Most graph- and role-based frameworks add only a thin layer, from
OpenAgents ($527$\,ms) and LangGraph ($548$\,ms) through CrewAI ($741$\,ms),
Agno ($902$\,ms), AutoGen ($964$\,ms), and the OpenAI SDK ($1{,}275$\,ms).
Concordia is the clear outlier at $30.6$\,s, roughly $60\times$ the baseline,
with throughput collapsing to $0.12$~req/s and output reaching
$\sim$100{,}000 characters against $\sim$15 elsewhere.

\revone{R1O2}{A trivial query issues identical model work everywhere, so cost
differences originate in framework machinery. We decompose wall-clock time at
the API boundary into LLM time and \emph{framework-residual}
time (Table~\ref{tab:overhead-decomp}). Residual stays under $21$\,ms and below
$2.2\%$ of latency for nearly all graph- and role-based frameworks, a light
wrapper over one model call. Agno is the exception at $39.3\%$. Concordia
inverts this, at $28.6$\,s and $91.5\%$ of latency, with $\sim$3 calls per query.
Prompt size does not explain it. \revthree{R3W2}{OpenHands sends as large a
prompt yet spends $12$\,ms residual.} The cost is Concordia's game-master
mediation, which routes action resolution and narrative updates through the
model, plus persistent loops and state serialization. The skeleton realizes the
same paradigm without that machinery, and its residual is $13.7$\,ms
($97.8\%$ LLM), three orders of magnitude below Concordia. Environment mediation
therefore costs milliseconds, and Concordia's overhead is implementation, not
paradigm (RQ5).} \revtwo{R2O4}{We report Concordia as deployed, because those
numbers describe the real cost of using it.} \revtwo{R2O3}{The light-wrapper
result spans six frameworks and is paradigm-level. Concordia's is
framework-level, and the skeleton supplies the GABM
reference.}\shorten

\subsubsection{Memory Architecture Effects}
\label{subsec:memoryagentbench}

We evaluate how memory architecture affects recall, in-session learning,
long-range reasoning, knowledge revision, and runtime scalability. All
frameworks are accessed through a common
interface (Section~\ref{subsec:bench_memory}) with identical chunking, ingestion
order, and metrics, and all experiments use the same backend model
(\texttt{gpt-oss-20b} on Groq\footnote{Although \texttt{gpt-oss-20b} supports
context windows up to 130K tokens, large accumulated contexts exceed
tokens-per-minute limits (e.g., 250K TPM), making extreme accumulation
impractical for evaluation.}). Observed differences reflect memory execution
semantics rather than model capability. \revthree{R3W1}{We verify across
capability with a Claude-Haiku-4.5 sensitivity row rather than a full frontier
sweep, because the suite is long-context and repeating it at frontier pricing is
prohibitive.\footnote{The suite costs roughly \$300 at \texttt{gpt-oss-20b}
pricing; frontier pricing raises this by one to two orders of magnitude.}}

\begin{table}[t]
\vspace{-2mm}
\centering
\footnotesize
\renewcommand{\arraystretch}{0.85}
\setlength{\tabcolsep}{3pt}
\caption{\revone{R1O2}{API-boundary cost decomposition on the trivial workload (50 trials, \emph{gpt-4o-mini}), ordered by residual $=$ total latency $-$ LLM time. The skeleton against Concordia separates GABM \emph{paradigm} cost from Concordia's \emph{implementation} cost. $^\dagger$out-of-taxonomy reference.}\revtwo{R2O4}{}}
\vspace{-4mm}
\label{tab:overhead-decomp}
\begin{tabular}{lrrrrr}
\toprule
Framework & p50 (ms) & Residual (ms) & Residual \% & Calls & Input tok. \\
\midrule
Direct LLM          & 495     & --      & --     & 1.0  & 30 \\
OpenAgents          & 527     & 0.06    & $<$0.1 & 1.0  & 14 \\
LangGraph           & 548     & 1.7     & 0.2    & 1.0  & 14 \\
AutoGen             & 964     & 5.9     & 0.6    & 1.0  & 26 \\
\revthree{R3W2}{OpenHands$^\dagger$} & 657 & 12.1 & 1.7 & 1.0 & 2{,}914 \\
\revone{R1O2}{GABM skeleton} & 522 & \textbf{13.7} & 2.2 & 1.0 & 46 \\
CrewAI              & 741     & 16.5    & 2.1    & 1.0  & 168 \\
OpenAI SDK          & 1{,}275 & 20.8    & 1.5    & 1.0  & 25 \\
Agno                & 902     & 365     & \textbf{39.3} & 1.0 & 14 \\
\revone{R1O2}{Concordia}\revtwo{R2O4}{} & 30{,}564 & \textbf{28{,}643} & \textbf{91.5} & \textbf{3.0} & 3{,}150 \\
\bottomrule
\end{tabular}
\vspace{-5mm}
\end{table}

\setlength{\tabcolsep}{4pt}
\begin{table*}[t]
\vspace{-3mm}
\centering
\caption{Memory architecture results on MemoryAgentBench. Three architectures are compared: retrieval-centric (CrewAI, LangGraph vector DB only), accumulation-based (Agno, OpenAI SDK), and hybrid (LangGraph with context window $W$). Scores are averaged per subtask across sessions. Category scores (AR, TTL, LRU, SF) are subtask means, and Overall is their average. Bold shaded cells indicate global maxima. Underlined values denote the best within each architecture.}
\vspace{-4mm}
\renewcommand{\arraystretch}{0.9}
\label{tab:memorybench-results}
\scriptsize
\begin{tabular}{
    l|
    *{5}{c}|  
    *{3}{c}|  
    *{3}{c}|  
    *{3}{c}|  
    c         
}
\toprule
\textbf{Agent Framework} &
\multicolumn{5}{c|}{\textbf{Accurate Retrieval (AR)}} &
\multicolumn{3}{c|}{\textbf{Test-Time Learning (TTL)}} &
\multicolumn{3}{c|}{\textbf{Long-Range Understanding (LRU)}} &
\multicolumn{3}{c|}{\textbf{Selective Forgetting (SF)}} &
\textbf{Overall} \\
\midrule
& SH-QA & MH-QA & LME(S*) & EventQA & Avg. &
  MCC & \ \ \  Recom.  \ \ \ & Avg. &
  \ \ \ \ Summ.\ \ \ \  & \ \ \ \ DetQA  \ \ \ \ \ \ \  & Avg. &
   \ \ \ FC-SH  \ \ \  &  \ \ \ FC-MH  \ \ \  & Avg. &
  Score \\
  
\midrule

\multicolumn{16}{c}{\cellcolor{gray!30}\textbf{\emph{Retrieval-centric} Memories (LLM = gpt-oss-20b)}} \\

Crewai & 14.0 & 22.0 & 3.0 & 48.7 & 21.9 & 5.6 & \underline{12.2} & 8.9 & 0.0 & 19.7 & 9.9 & 26.2 & 0.8 & 13.5 & 13.5 \\
LangGraph (Vector DB only) & \underline{14.0} & \underline{24.0} & \underline{31.3} & \underline{63.3} & \underline{33.2} & \underline{22.8} & 0.0 & \underline{11.4} & \underline{20.0} &\underline{40.8} & \cellcolor{YellowGreen!30}{\textbf{\underline{30.4}}} & \underline{36.0} & \underline{4.5} & \cellcolor{YellowGreen!30}{\textbf{\underline{20.2}}} & \cellcolor{YellowGreen!30}{\textbf{23.8}} \\

\multicolumn{16}{c}{\cellcolor{gray!30}\textbf{\emph{Accumulation-based} Memories (LLM = gpt-oss-20b)}} \\

Agno & 12.0 & 19.0 & 2.0 & 43.2 & 19.1 & 0.6 & 15.1 & 7.8 & 0.0 & 18.3 & 9.2 & 27.0 & 0.8 & 13.9 & 12.5 \\

OpenAI SDK ($W=50$) & 16.0 & 16.0 & 1.7 & 0.0 & 8.4 & 1.4 & 0.9 & 1.2 & 0.0 & 0.0 & 0.0 & 28.5 & \underline{1.0} & \underline{14.8} & 6.1 \\

OpenAI SDK ($W=512$) & 8.0 & 11.0 & 1.0 & 44.1 & 16.0 & 0.6 & 17.2 & 8.9 & 0.0 & 28.2 & 14.1 & \underline{28.7} & 0.8 & 14.8 & 13.4 \\

OpenAI SDK ($W=1024$) & \underline{40.0} & 33.0 & 7.7 & 48.5 & 32.3 & 3.6 & 22.1 & 12.8 & 0.0 & 25.4 & 12.7 & 0.0 & 0.0 & 0.0 & 14.5 \\

OpenAI SDK ($W=2048$) & 32.0 & 32.0 & 18.3 & \underline{49.9} & 33.1 & 10.0 & 20.9 & 15.5 & 0.0 & 33.8 & 16.9 & 0.2 & 0.0 & 0.1 & 16.4 \\

OpenAI SDK ($W=4096$) & 36.0 & \underline{35.0} & 14.0 & 48.5 & 33.4 & 16.2 & \underline{22.7} & 19.4 & 0.0 & 29.6 & 14.8 & 0.5 & 0.0 & 0.2 & 17.0 \\

OpenAI SDK ($W=8192$) & 34.0 & 34.0 & \underline{23.0} & 44.7 & \underline{33.9} & \underline{34.0} & 7.3 & \underline{20.7} & \underline{10.0} & \underline{45.1} & \underline{27.5} & 0.8 & 0.0 & 0.4 & 20.6 \\

\multicolumn{16}{c}{\cellcolor{gray!30}\textbf{\emph{Hybrid Retrieval–Accumulation} Memories (LLM = gpt-oss-20b)}} \\

LangGraph ($W=50$) & 35.0 & 35.0 & 41.7 & 3.7 & 28.8 & \underline{38.6} & 3.9 & 21.3 & 0.0 & 1.4 & 0.7 & \underline{0.2} & 0.2 & 0.2 & 12.8 \\

LangGraph ($W=512$) & 36.0 & 34.0 & \underline{44.0} & \underline{65.6} & \cellcolor{YellowGreen!30}{\textbf{\underline{44.9}}} & 35.6 & 12.8 & 24.2 & 0.0 & 35.2 & 17.6 & 0.0 & 0.0 & 0.0 & 21.7 \\

LangGraph ($W=1024$)  & 34.0 & \underline{35.0} & 41.3 & 64.2 & 43.6 & 38.0 & \underline{19.8} & \cellcolor{YellowGreen!30}{\textbf{\underline{28.9}}} & 0.0 & 40.8 & 20.4 & 0.0 & 0.2 & 0.1 & \underline{23.3} \\

LangGraph ($W=2048$)  & \underline{37.0} & 30.0 & 39.3 & 63.7 & 42.5 & 34.6 & 19.4 & 27.0 & 0.0 & 40.8 & 20.4 & 0.0 & \underline{0.5} & \underline{0.2} & 22.5 \\

LangGraph ($W=4096$)  & 33.0 & 34.0 & 36.7 & 60.4 & 41.0 & 32.2 & 14.6 & 23.4 & 0.0 & 29.6 & 14.8 & 0.0 & 0.0 & 0.0 & 19.8 \\

LangGraph ($W=8192$)  & 31.0 & 35.0 & 41.7 & 59.6 & 41.8 & 32.8 & 18.7 & 25.7 & \underline{10.0} & \underline{40.8} & \underline{25.4} & 0.0 & 0.2 & 0.1 & 23.3 \\

\multicolumn{16}{c}{\cellcolor{gray!30}\textbf{\emph{Model Sensitivity Check (LLM = Claude-Haiku-4.5)}}} \\

LangGraph (Vector DB only) & 9.0 & 8.0 & 35.3 & 82.5 & 33.7 & 0.8 & 51.6 & 26.2 & 40.0 & 53.5 & 46.8 & 29.0 & 4.0 & 16.5 & 30.8 \\

\bottomrule

\end{tabular}
\vspace{-3mm}
\end{table*}

Table~\ref{tab:memorybench-results} reveals three dominant memory architectures:
\emph{retrieval-centric} memory with persistent semantic stores,
\emph{accumulation-based} memory that replays recent interactions in the prompt,
and \emph{hybrid} memory combining retrieval with bounded short-term
accumulation. CrewAI is retrieval-centric. Agno and the OpenAI SDK are
accumulation-based. LangGraph adopts a hybrid design. We evaluate LangGraph in
retrieval-only and hybrid modes across context windows ($W$) and vary $W$ for
the OpenAI SDK to characterize accumulation scaling. Concordia and OpenAgents
are excluded for narrative simulation semantics and absent native memory
abstractions.\shorten

\noindent\textbf{Accuracy and Reasoning Effects.}
Hybrid retrieval--accumulation performs best on Accurate Retrieval (AR).
LangGraph ($W{=}512$) reaches \emph{44.9}, above pure retrieval (\emph{33.2})
and accumulation-based Agno (\emph{19.1}). Retrieval dominates because it
isolates relevant evidence from stored history, and accumulation degrades
because unfiltered context dilutes signal (RQ1a). 
Accumulation-only memory rises to \emph{32.3} for the OpenAI SDK at $W{=}1024$
and stays almost flat as context grows. Reliable factual recall depends on
architectural retrieval, not larger context windows.\shorten

Long-Range Understanding (LRU) shows a similar pattern. Retrieval-only LangGraph
achieves the highest score (\emph{30.4}). Accumulation-based designs approach it
only at very large context windows (OpenAI SDK at $W{=}8192$: \emph{27.5}), and
at higher cost. Hybrid designs perform worse on LRU, because short-term
accumulation interferes with retrieved narratives and weakens temporal
coherence (RQ1b).

\noindent\textbf{In-Session Learning and Knowledge Revision.}
Test-Time Learning (TTL) shows the opposite trend. Accumulation-based designs
improve steadily as history grows, with the OpenAI SDK rising from \emph{1.2}
at $W{=}50$ to \emph{20.7} at $W{=}8192$. Hybrid designs outperform pure
accumulation, and LangGraph ($W{=}1024$) achieves the highest TTL
score (\emph{28.9}) by anchoring learning to retrieved signals rather than
prompt growth. Retrieval-only memory performs poorly, so adaptation requires
accumulation but benefits from filtering (RQ1c).

Selective Forgetting (SF) remains largely unsupported. Retrieval-centric designs
partially support single-hop forgetting, with retrieval-only LangGraph reaching
\emph{20.2}, but multi-hop forgetting stays below \emph{5.0} across all
architectures. Accumulation-based designs perform worse even at large context
sizes, and hybrid designs collapse as outdated information persists in retrieval
stores and accumulated context at once. \revthree{R3W4}{Every evaluated
framework-native architecture treats memory as append-only.} Controlled
knowledge revision requires explicit state update mechanisms, such as revise or
remove, that no evaluated framework provides (RQ1d).

\noindent\textbf{Effect of Model Capability.}
We evaluate LangGraph (Vector DB only) using Claude-Haiku-4.5
(Table~\ref{tab:memorybench-results}, last row). AR stays stable (33.2 vs.\
33.7), and the stronger LLM under retrieval-only design still does not reach the
hybrid score of 44.9 with a weaker model, so architecture is the
bottleneck (RQ1a). TTL improves, yet accumulation with a weaker LLM still
outperforms retrieval-only with a stronger one, preserving the architectural
ranking (RQ1c). LRU improves (30.4 to 46.8), so architecture and model
capability contribute jointly. SF declines slightly, so knowledge revision
remains unsupported regardless of model strength (RQ1d). \revthree{R3W1}{Model
capability affects absolute scores but does not alter architectural rankings or
change the answer to any question.}

\begin{figure}[t]
\centering
\vspace{2mm}
\setlength{\abovecaptionskip}{2pt}
\setlength{\belowcaptionskip}{-6pt}
\begin{tikzpicture}
\begin{groupplot}[
  group style={group size=2 by 1, horizontal sep=0.3cm},
  width=0.52\linewidth,
  height=0.38\linewidth,
  symbolic x coords={50,512,1024,2048,4096,8192},
  xtick=data,
  xlabel={Context Window (tokens)},
  ylabel={Total Runtime (s)},
  grid=both,
  grid style={dashed,gray!30},
  tick label style={font=\scriptsize},
  label style={font=\scriptsize},
  title style={font=\scriptsize, yshift=-7}, 
  legend style={
  font=\scriptsize,
  draw=none,
  legend image code/.code={
    \draw[mark size=0.8pt]
      plot coordinates {(0cm,0cm) (0.2cm,0cm)};
  }
}
]
\nextgroupplot[
  yshift=10pt,
  title={LangGraph},
  legend style={at={($(0,0)+(-1cm,-1cm)$)},legend columns=1,fill=none,draw=black,anchor=center,align=center},
  legend to name=fred
]

\addplot+[mark=*, mark size=1.2pt, thick, blue]
  coordinates {(50,4236.5) (512,4538.9) (1024,4495.4) (2048,5093.6) (4096,5817.5) (8192,6705.5)};
\addlegendentry{AR}

\addplot+[mark=*, mark size=1.2pt, thick, orange]
  coordinates {(50,2247.6) (512,2336.1) (1024,2487.4) (2048,2607.0) (4096,2896.3) (8192,3323.9)};
\addlegendentry{TTL}

\addplot+[mark=*, mark size=1.2pt, thick, green!60!black]
  coordinates {(50,529.3) (512,523.5) (1024,507.1) (2048,548.7) (4096,531.4) (8192,541.3)};
\addlegendentry{LRU}

\addplot+[mark=*, mark size=1.2pt, thick, red]
  coordinates {(50,1861.8) (512,2043.3) (1024,2049.5) (2048,2094.1) (4096,2086.5) (8192,2099.3)};
\addlegendentry{SF}

\nextgroupplot[
  title={OpenAI SDK},
  ylabel={},
  yticklabels={},
  ytick=\empty,
]

\addplot+[mark=*, mark size=1.2pt, thick, blue]
  coordinates {(50,5475.7) (512,5192.0) (1024,4829.4) (2048,5266.9) (4096,5838.1) (8192,6955.1)};

\addplot+[mark=*, mark size=1.2pt, thick, orange]
  coordinates {(50,2175.9) (512,2557.4) (1024,3064.1) (2048,2807.4) (4096,2733.7) (8192,2873.4)};

\addplot+[mark=*, mark size=1.2pt, thick, green!60!black]
  coordinates {(50,1650.0) (512,1598.0) (1024,1479.2) (2048,1609.5) (4096,1805.3) (8192,2316.5)};

\addplot+[mark=*, mark size=1.2pt, thick, red]
  coordinates {(50,2450.6) (512,1846.7) (1024,1832.3) (2048,1978.4) (4096,1718.5) (8192,3382.1)};

\end{groupplot}
\node[
  anchor=west,
  xshift=6pt
] at (group c2r1.east) {\ref{fred}};
\end{tikzpicture}
\caption{Runtime scaling with context window for LangGraph (hybrid retrieval--accumulation, left) and OpenAI SDK (pure accumulation, right) across four MemoryAgentBench competencies. Accumulation drives steep growth on high-query tasks (AR, TTL), while retrieval stays stable.}
\label{fig:runtime-vs-context}
\vspace{-2mm}
\end{figure}

\noindent\textbf{Runtime and Scalability Effects.}
Figure~\ref{fig:runtime-vs-context} reports runtime against context window for
LangGraph (hybrid) and the OpenAI SDK (pure accumulation). High-query tasks such
as AR and TTL grow steeply under accumulation, where full histories are replayed
for each query. LRU stays nearly flat for LangGraph because sessions carry few
queries, and SF grows moderately. Retrieval avoids repeated processing, so
scalability is governed by memory architecture rather than by the LLM's nominal
context capacity. \revtwo{R2O3}{Retrieval-centric and accumulation-based results
each span two independent frameworks and are architecture-level. The hybrid
result rests on LangGraph alone and is framework-level.}

\subsubsection{\revone{R1O3}{Planning Interface Effects}}
\label{subsec:planning_results}
We evaluate how the \emph{planning interface} affects reasoning accuracy,
robustness, and runtime, separating architectural constraints from the LLM's
planning ability. \revone{R1O3}{The same model produces the plan under Crew-Plan
and Direct-LLM-Plan, so the comparison isolates the interface.} Using the
MAFBench planning pipeline (Section~\ref{subsec:bench_planning}), we fix the
benchmarks (GSM8K~\cite{cobbe2021training},
CSQA~\cite{talmor-etal-2019-commonsenseqa}, MATH~\cite{hendrycks2measuring}),
prompts, decoding settings, and scoring, and vary only the interface: \emph{NoPlan}
(direct answer), \emph{Crew-Plan} (schema-constrained two-stage planning in
CrewAI~\cite{crewai2025}), and \emph{Direct-LLM-Plan} (free-form plan text
injected before execution). For MATH we evaluate a 100-problem subset preserving
the original complexity distribution.

Table~\ref{tab:reasoning-accuracy} shows that schema-constrained planning
reduces accuracy across datasets and models, but free-form planning preserves or
improves it for weaker ones. Under Crew-Plan, Qwen-7B drops from \emph{13.0\%}
to \emph{3.4\%} on GSM8K and from \emph{69.9\%} to \emph{37.6\%} on CSQA, and
GPT-OSS-20B drops from \emph{80.0\%} to \emph{48.0\%} on MATH-100.
Direct-LLM-Plan instead lifts Qwen-7B to \emph{28.7\%} on GSM8K and Llama-3.1-8B
from \emph{65.6\%} to \emph{71.9\%}. Tables~\ref{tab:reasoning-failures}
and~\ref{tab:reasoning-time-mult} identify the architectural cause. Crew-Plan
produces large formatting failure rates, where the model violates the
framework's required schema and the plan cannot be parsed even when the
reasoning is correct. This reaches \emph{84.7\%} on GSM8K for Qwen-7B. Crew-Plan
also adds an extra call with schema generation, validation, and parsing, raising
runtime \emph{7.4$\times$} on GSM8K and \emph{18.5$\times$} on CSQA for
GPT-OSS-20B, and exceeding \emph{31$\times$} for Llama-3.1-8B on CSQA.
Direct-LLM-Plan yields \emph{0.0\%} formatting failures across all models, with
smaller slowdowns (\emph{0.8$\times$--6.6$\times$}).

\revthree{R3W1}{Capability bounds the failure mechanism but not the cost.
GPT-5.6-Luna, GPT-5.6-Terra, and Claude-Opus-5 never violate the schema, so
parse failure is a weak-model phenomenon. Eliciting a plan does not help them
either. Across all three benchmarks and both interfaces their accuracy remains
almost flat, yet overhead persists at \emph{2.5$\times$--4.0$\times$} for
GPT-5.6 and \emph{9.2$\times$--15.0$\times$} for Opus-5, whose longer plans make
its absolute cost the largest measured. The interface therefore carries two
independent costs. Parse failure scales inversely with capability and vanishes
at the frontier. Plan injection and the extra call add runtime and token costs
without benefit when tasks fit the model's context window. More complex tasks,
such as software engineering, are future work.} These results answer RQ2. Below
the frontier, \revone{R1O3}{formatting failures rather than reasoning errors are
the primary cause of accuracy loss under the schema-constrained interface}. At
the frontier the schema is satisfied and the residual cost is the plan itself.
\revtwo{R2O3}{CrewAI is the only evaluated framework with a runtime planning
interface, so this is framework-level. It holds across ten models spanning two
orders of magnitude in capability, so it is not model-specific.}

\setlength{\tabcolsep}{1.7pt}
\begin{table}[t]
\centering
\renewcommand{\arraystretch}{.75}
\scriptsize
\caption{Accuracy (\%) under three planning interfaces: \xmark\, no plan, \checkmark(Crew)\, schema-constrained, \checkmark(LLM)\, free-form. Schema-constrained planning reduces accuracy for weaker models through formatting failures.
\revone{R1O3}{Both planning modes use the same model to generate the plan.}}
\vspace{-3mm}
\renewcommand{\arraystretch}{0.65}
\label{tab:reasoning-accuracy}
\begin{tabular}{l|ccc|ccc|ccc}
\toprule
\multirow{2}{*}{\textbf{LLM}} & \multicolumn{3}{c}{\textbf{GSM8K}~\cite{cobbe2021training}} & \multicolumn{3}{c}{\textbf{CSQA}~\cite{talmor-etal-2019-commonsenseqa}} & \multicolumn{3}{c}{\textbf{MATH-100}~\cite{hendrycks2measuring}} \\
\cmidrule(lr){2-4} \cmidrule(lr){5-7} \cmidrule(lr){8-10}
 & \xmark & \checkmark (crew) & \checkmark (LLM) & \xmark & \checkmark (crew) & \checkmark (LLM)& \xmark & \checkmark (crew) & \checkmark (LLM) \\
\midrule
\rowcolor{gray!20} \multicolumn{10}{c}{\textbf{Local Models}} \\
GPT-OSS-20B~\cite{openai2025gptoss} & 94.6 & \accChange{94.6}{91.6} & \accChange{94.6}{87.6} & 81.4 & \accChange{81.4}{81.2} & \accChange{81.4}{82.1} & 80.0 & \accChange{80.0}{48.0} & \accChange{80.0}{76.0} \\
Phi-4-14B~\cite{abdin2024phi4} & 93.3 & \accChange{93.3}{86.7} & \accChange{93.3}{83.5} & 81.7 & \accChange{81.7}{82.0} & \accChange{81.7}{74.6} & 72.0 & \accChange{72.0}{50.0} & \accChange{72.0}{69.0} \\
Llama-3.1-8B~\cite{dubey2024llama3} & 65.6 & \accChange{65.6}{60.3} & \accChange{65.6}{71.9} & 70.1 & \accChange{70.1}{65.4} & \accChange{70.1}{65.4} & 25.0 & \accChange{25.0}{24.0} & \accChange{25.0}{37.0} \\
Qwen-7B~\cite{bai2024qwenvl} & 13.0 & \accChange{13.0}{3.4} & \accChange{13.0}{28.7} & 69.9 & \accChange{69.9}{37.6} & \accChange{69.9}{74.0} & 9.0 & \accChange{9.0}{4.0} & \accChange{9.0}{13.0} \\

DeepSeek-7B~\cite{deepseek2024v3} & 27.0 & \accChange{27.0}{18.0} & \accChange{27.0}{23.5} & 46.3 & \accChange{46.3}{31.2} & \accChange{46.3}{46.1} & 9.0 & \accChange{9.0}{3.0} & \accChange{9.0}{4.0} \\
\rowcolor{gray!20} \multicolumn{10}{c}{\textbf{Remote Models}} \\
GPT-4.1~\cite{openai2024gpt4} & 94.2 & \accChange{94.2}{80.7} & \accChange{94.2}{78.2} & 87.1 & \accChange{87.1}{87.3} & \accChange{87.1}{87.0} & 86.0 & \accChange{86.0}{60.0} & \accChange{86.0}{84.0} \\
GPT-4o-Mini~\cite{openai2024gpt4} & 86.5 & \accChange{86.5}{90.8} & \accChange{86.5}{84.2} & 81.8 & \accChange{81.8}{81.3} & \accChange{81.8}{83.1} & 60.0 & \accChange{60.0}{53.0} & \accChange{60.0}{59.0} \\

\revthree{R3W1}{GPT-5.6-Luna} & 96.5 & \accChange{96.5}{96.7} & \accChange{96.5}{96.7} & 87.7 & \accChange{87.7}{86.4} & \accChange{87.7}{86.1} & 90.0 & 
\accChange{90.0}{90.0} 
& \accChange{90.0}{89.0} \\

\revthree{R3W1}{GPT-5.6-Terra} & 96.4 & \accChange{96.4}{95.8} & \accChange{96.4}{92.6} & 87.6 & \accChange{87.6}{87.2} & \accChange{87.6}{86.0} & 94.0 & \accChange{94.0}{93.0} & \accChange{94.0}{94.0} \\

\revthree{R3W1}{Claude-Opus-5} & 97.7 & \accChange{97.7}{97.1} & \accChange{97.7}{96.8} & 90.7 & \accChange{90.7}{87.8} & \accChange{90.7}{89.3} & 95.0 & \accChange{95.0}{95.0} & \accChange{95.0}{95.0} \\

\bottomrule
\end{tabular}
\vspace{-2mm}
\end{table}

\setlength{\tabcolsep}{3.pt}
\begin{table}[t]
\centering
\scriptsize
\caption{Percentage of questions where the LLM output violated the required schema, under Crew-Plan and Direct-LLM-Plan. Crew-Plan failures reach 84.7\% for weaker models. Direct-LLM-Plan produces zero violations across all models.}
\vspace{-3mm}
\renewcommand{\arraystretch}{0.85}
\label{tab:reasoning-failures}
\begin{tabular}{l|cc|cc|cc}
\toprule
\multirow{2}{*}{\textbf{LLM}} & \multicolumn{2}{c}{\textbf{GSM8K}} & \multicolumn{2}{c}{\textbf{CSQA}} & \multicolumn{2}{c}{\textbf{MATH-100}} \\
\cmidrule(lr){2-3} \cmidrule(lr){4-5} \cmidrule(lr){6-7}
 & \checkmark (CrewAI) & \checkmark (LLM) & \checkmark (CrewAI) & \checkmark (LLM) & \checkmark (CrewAI) & \checkmark (LLM) \\
\midrule
\rowcolor{gray!20} \multicolumn{7}{c}{\textbf{Local Models}} \\
GPT-OSS-20B & 2.4\% & 0.0\% & 1.3\% & 0.0\% & \textbf{49.0\%} & 0.0\% \\
Phi-4-14B & 6.1\% & 0.0\% & 0.5\% & 0.0\% & 21.0\% & 0.0\% \\
Llama-3.1-8B & 10.5\% & 0.0\% & 4.8\% & 0.0\% & 19.0\% & 0.0\% \\
Qwen-7B & \textbf{84.7\%} & 0.0\% & \textbf{41.0\%} & 0.0\% & \textbf{70.0\%} & 0.0\% \\
DeepSeek-7B & 25.5\% & 0.0\% & 31.0\% & 0.0\% & \textbf{40.0\%} & 0.0\% \\
\rowcolor{gray!20} \multicolumn{7}{c}{\textbf{Remote Models}} \\
GPT-4.1 & 0.0\% & 0.0\% & 0.0\% & 0.0\% & 0.0\% & 0.0\% \\
GPT-4o-Mini & 0.0\% & 0.0\% & 0.0\% & 0.0\% & 0.0\% & 0.0\% \\
\revthree{R3W1}{GPT-5.6-Luna} & 0.0\% & 0.0\% & 0.0\% & 0.0\% & 0.0\% & 0.0\% \\
\revthree{R3W1}{GPT-5.6-Terra} & 0.0\% & 0.0\% & 0.0\% & 0.0\% & 0.0\% & 0.0\% \\

\revthree{R3W1}{Claude-Opus-5} & 0.0\% & 0.0\% & 0.0\% & 0.0\% & 0.0\% & 0.0\% \\

\bottomrule
\end{tabular}
\vspace{-3mm}
\end{table}

\setlength{\tabcolsep}{4.pt}
\begin{table}[t]
\centering
\scriptsize
\caption{Runtime overhead relative to NoPlan under Crew-Plan and Direct-LLM-Plan. Crew-Plan increases runtime by up to 31$\times$. Direct-LLM-Plan adds modest overhead (0.8$\times$--6.6$\times$). Lower values indicate less overhead. \revthree{R3W1}{Overhead persists at frontier capability, and Opus-5 pays the largest, since it generates longer plans.}}
\vspace{-3mm}
\renewcommand{\arraystretch}{0.75}
\label{tab:reasoning-time-mult}
\begin{tabular}{l|cc|cc|cc}
\toprule
\multirow{2}{*}{\textbf{LLM}} & \multicolumn{2}{c}{\textbf{GSM8K}} & \multicolumn{2}{c}{\textbf{CSQA}} & \multicolumn{2}{c}{\textbf{MATH}} \\
\cmidrule(lr){2-3} \cmidrule(lr){4-5} \cmidrule(lr){6-7}
 & $\Delta$ CrewAI & $\Delta$ LLM & $\Delta$ CrewAI & $\Delta$ LLM & $\Delta$ CrewAI & $\Delta$ LLM \\
\midrule
\rowcolor{gray!20} \multicolumn{7}{c}{\textbf{Local Models}} \\
GPT-OSS-20B & $\times$ 7.4 & $\times$ 2.3 & $\times$ 18.51 & $\times$ 1.7 & $\times$ 11.43 & $\times$ 1.7 \\
Phi-4-14B & $\times$ 3.5 & $\times$ 1.4 & $\times$ 4.1 & $\times$ 1.6 & $\times$ 3.0 & $\times$ 1.7 \\
Llama-3.1-8B & $\times$ 6.3 & $\times$ 1.3 & $\times$ 31.34 & $\times$ 4.2 & $\times$ 4.4 & $\times$ 1.3 \\
Qwen-7B & $\times$ 2.2 & $\times$ 2.5 & $\times$ 18.16 & $\times$ 0.9 & $\times$ 2.9 & $\times$ 1.8 \\
DeepSeek-7B & $\times$ 3.1 & $\times$ 2.6 & $\times$ 13.29 & $\times$ 6.6 & $\times$ 2.8 & $\times$ 1.2 \\
\rowcolor{gray!20} \multicolumn{7}{c}{\textbf{Remote Models}} \\
GPT-4.1 & $\times$ 2.4 & $\times$ 1.5 & $\times$ 2.9 & $\times$ 1.1 & $\times$ 1.2 & $\times$ 1.1 \\
GPT-4o-Mini & $\times$ 1.5 & $\times$ 1.4 & $\times$ 4.8 & $\times$ 2.3 & $\times$ 1.8 & $\times$ 1.4 \\
\revthree{R3W1}{GPT-5.6-Luna} & $\times$ 2.6 & $\times$ 1.6 & $\times$ 4.0 & $\times$ 1.5 & $\times$ 2.5 & $\times$ 1.2 \\
\revthree{R3W1}{GPT-5.6-Terra} & $\times$ 2.9 & $\times$ 1.5 & $\times$ 3.2 & $\times$ 0.8 & $\times$ 2.5 & $\times$ 1.3 \\

\revthree{R3W1}{Claude-Opus-5} & $\times$ 15.00 & $\times$ 2.7 & $\times$ 14.57 & $\times$ 2.6 & $\times$ 9.2 & $\times$ 2.3 \\

\bottomrule
\end{tabular}
\vspace{-3mm}
\end{table}

\subsubsection{Agent Specialization}
\label{sec:specialization}

\begin{table}[t]
\centering
\scriptsize
\setlength{\tabcolsep}{2.pt}
\vspace{-2mm}
\caption{Specialization under four conditioning strategies (GPT-4o-mini).
\revone{R1O4}{Identity conditioning gives no gain over \emph{No Role}.
Capability conditioning helps only when it supplies procedure, so the
dataset-profile tool does not reproduce the expert-guided result.}}
\label{tab:specialaiztion_result}
\vspace{-3mm}
\renewcommand{\arraystretch}{0.9}
\begin{tabular}{l|c|ccc|ccc|ccc|ccc}
\toprule
\textbf{Category} & \textbf{Regres.} & \multicolumn{3}{c|}{\textbf{Binary Class}} & \multicolumn{9}{c}{\textbf{Multiclass}} \\
\midrule
\multirow{2}{*}{\textbf{Agent Role}} 
& \multicolumn{1}{c|}{\textbf{Utility}} 
& \multicolumn{3}{c|}{\textbf{WiFi}} 
& \multicolumn{3}{c|}{\textbf{EU-IT}} 
& \multicolumn{3}{c|}{\textbf{Yelp}} 
& \multicolumn{3}{c}{\textbf{Volkert}} \\ 
\cmidrule(lr){2-2} \cmidrule(lr){3-5} \cmidrule(lr){6-8} \cmidrule(lr){9-11} \cmidrule(lr){12-14}
& \textbf{MAE} 
& \textbf{P} & \textbf{R} & \textbf{F1} 
& \textbf{P} & \textbf{R} & \textbf{F1} 
& \textbf{P} & \textbf{R} & \textbf{F1} 
& \textbf{P} & \textbf{R} & \textbf{F1} \\
\midrule
\multicolumn{14}{c}{\cellcolor{gray!20}\textbf{\revone{R1O4}{Role-as-identity:} role-based prompting}} \\

No Role        & 0.067 & 40.2 & 45.0 & 37.3  
& 32.0 & 38.0 & 35.0 
& 41.9 & 44.1 & 41.9 
& 67.3 & 66.5 & 65.3 \\

Data Scientist & 0.067 & 40.2 & 45.0 & 37.3 
& 32.3 & 37.9 & 34.8 
& 41.9 & 44.1 & 41.9 
& 67.3 & 66.5 & 65.3 \\

Researcher     & 0.067 & 40.2 & 45.0 & 37.3 
& 32.3 & 37.9 & 34.8 
& 42.1 & 44.0 & 41.9 
& 67.3 & 67.0 & 65.8 \\

Data Analyst   & 0.067 & 40.2 & 45.0 & 37.3 
& 33.3 & 37.9 & 34.8 
& 42.1 & 44.0 & 41.9
& 67.0 & 67.0 & 65.0 \\

Engineer       & 0.067 & 40.2 & 45.0 & 37.3 
& 32.0 & 38.0 & 35.0 
& 41.9 & 44.1 & 41.9 
& 67.3 & 66.5 & 65.3 \\

\multicolumn{14}{c}{\cellcolor{gray!20}\textbf{\revone{R1O4}{Role label with} self-generated plan}} \\

No Role        & 0.067 & 40.0 & 45.0 & 37.0 
& 32.0 & 38.0 & 35.0 
& 42.0 & 44.1 & 42.0
& 67.0 & 67.0 & 65.0 \\

Data Scientist & 0.067 & 40.0 & 45.0 & 37.0 
& 32.0 & 38.0 & 35.0 
& 42.0 & 44.1 & 42.0 
& 67.0 & 67.0 & 66.0 \\

Researcher     & 0.067 & 40.0 & 45.0 & 37.0
& 32.0 & 38.0 & 35.0 
& 42.0 & 44.0 & 42.0 
& 67.0 & 67.0 & 65.0 \\

Data Analyst   & 0.067 & 40.0 & 45.0 & 37.0
& 32.0 & 38.0 & 35.0 
& 42.0 & 44.0 & 42.0 
& 67.0 & 67.0 & 65.0 \\

Engineer       & 0.067 & 40.0 & 45.0 & 37.0
& 32.0 & 38.0 & 35.0 
& 42.0 & 44.0 & 42.0 
& 67.0 & 67.0 & 65.0 \\

\multicolumn{14}{c}{\cellcolor{gray!20}\textbf{\revone{R1O4}{Role-as-capability:} expert-guided procedure}} \\

No Role        & 0.067 & 53.0 & 90.0 & 67.0 
& 33.6 & 41.4 & 36.8 
& 96.0 & 96.0 & 96.0 
& 94.8 & 94.7 & 94.6 \\

Data Scientist & 0.068 & 57.1 & 80.0 & 66.7 
& 37.3 & 44.8 & 40.3 
& 100.0 & 100.0 & 100.0 
& 95.0 & 95.0 & 95.0 \\

Researcher     & 0.068 & 53.0 & 90.0 & 67.0  
& 37.3 & 41.4 & 38.8 
& 100.0 & 100.0 & 100.0 
& 95.0 & 94.9 & 94.9 \\

Data Analyst   & 0.067 & 52.9 & 90.0 & 66.7 
& 37.3 & 41.4 & 38.8 
& 100.0 & 100.0 & 100.0 
& 94.7 & 94.6 & 94.5 \\

Engineer       & 0.068 & 50.0 & 90.0 & 64.3 
& 37.3 & 41.4 & 38.8 
& 100.0 & 100.0 & 100.0 
& 94.8 & 94.7 & 94.6 \\

\multicolumn{14}{c}{\cellcolor{gray!20}\textbf{\revone{R1O4}{Role-as-capability: dataset-profile tool}}} \\

No Role        & 0.0682 & 62.2 & 65.0 & 62.0 
& 36.8 & 28.3 & 25.6 
& 45.6 & 46.0 & 40.8 
& 66.5 & 66.1 & 64.6 \\

Data Scientist & 0.07 & 62.2 & 65.0 & 62.0 
& 24.9 & 35.6 & 24.9 
& 44.5 & 45.8 & 41.0 
& 67.3 & 66.9 & 65.7 \\

Researcher     & 0.065 & 73.3 & 84.6 & \textbf{78.6}
& 25.7 & 35.6 & 25.5 
& 44.2 & 45.7 & 41.0 
& 66.5 & 66.1 & 64.6 \\

Data Analyst   & 0.049 & 63.9 & 65.0 & 64.3 
& 33.9 & 39.2 & 32.7 
& 45.5 & 45.6 & 39.1 
& 67.5 & 66.7 & 65.4 \\

Engineer       & 0.068 & 68.8 & 84.6 & 75.9 
& 17.5 & 30.8 & \textbf{18.7}
& 46.1 & 45.5 & 38.9 
& 66.5 & 66.1 & 64.6 \\
\bottomrule
\end{tabular}
\vspace{-3mm}
\end{table}

We evaluate how specialization mechanisms shape domain-specific reasoning
behavior. \revone{R1O4}{A role can name an identity or bind a capability.} Using
the unified specialization pipeline (Section~\ref{subsec:bench_specialization}),
we fix the LLM backend (GPT-4o-mini), datasets from CatDB~\cite{catdb}, prompts,
task structure, and evaluation metrics, and vary only the conditioning strategy
across the four conditions of Table~\ref{tab:specialaiztion_result}.
\revone{R1O4}{A professional label and a label with self-generated planning
supply identity, binding nothing to the role. Expert-guided instructions and the
dataset-profile tool bind a capability. The tool grants data access with no
prescribed workflow, which isolates it from procedural knowledge.}

Identity conditioning gives nearly identical performance across all professional
labels, with no gain over the \emph{No Role} baseline, and self-generated
planning stabilizes behavior without improving accuracy. Labels fail because
they do not constrain the solution workflow. Expert-guided procedure does,
lifting F1 above $94$ on Yelp and Volkert against roughly $42$ and $65$ under
labels.\footnote{Example prompts are provided in the
\uline{\textcolor{blue}{\href{https://github.com/CoDS-GCS/MAFBench/blob/main/supplementary_materials/APPENDICIES.pdf}{supplementary materials}}}.}
\revone{R1O4}{The dataset-profile tool separates data access from procedure. It
helps where the obstacle is knowing the data, raising WiFi F1 from $37.3$ to
$78.6$, above the expert-guided range. It does not help where the obstacle is
method. Yelp and Volkert stay near their label values, and EU-IT falls to $18.7$
as the profile pushes the model toward features it cannot exploit. Bound
capability therefore aids specialization only when it constrains the
workflow (RQ3).} 
\revtwo{R2O3}{Conditioning is applied in the prompt,
independent of any framework, so this result is not tied to one
implementation.}

\setlength{\tabcolsep}{1.pt}  %
\vspace{-2mm}
\begin{table*}[t]
\centering
\renewcommand{\arraystretch}{.75}
\scriptsize
\caption{Coordination scalability across topology classes on \textsc{AGENTSNET}. Each entry reports performance for $n=4,8,16,50,100$ agents. All ($^*$) simulates GABM-style centralized communication using a fully connected graph. Runs with rounds $>40$ are excluded for sequential pipelines at large $n$. Only fully connected topology achieves consistent consensus\revthree{R3M1}{, answering RQ4}.}
\vspace{-4mm}
\renewcommand{\arraystretch}{0.85}
\label{tab:scalability-results}
\begin{tabular}{p{1.5cm}p{1.8cm}ccccc|ccccc|ccccc||ccccc|ccccc||ccccc}
\toprule
 & & \multicolumn{5}{c|}{\textbf{Graph-based (Small-World)}} 
   & \multicolumn{5}{c|}{\textbf{Graph-based (Scale-Free)}} 
   & \multicolumn{5}{c|}{\textbf{Graph-based (Delaunay)}} 
   & \multicolumn{5}{c|}{\textbf{Role-based (Sequential)}} 
   & \multicolumn{5}{c|}{\textbf{Role-based (Hierarchical)}} 
   & \multicolumn{5}{c}{\textbf{GABM-based (All)$^*$}} \\
\cmidrule(lr){3-7} \cmidrule(lr){8-12} \cmidrule(lr){13-17} \cmidrule(lr){18-22} \cmidrule(lr){23-27} \cmidrule(lr){28-32}
\textbf{Task} & \textbf{Metric / $n$} & $4$ & $8$ & $16$ & $50$ & $100$ & $4$ & $8$ & $16$ & $50$ & $100$ & $4$ & $8$ & $16$ & $50$ & $100$ & $4$ & $8$ & $16$ & $50$ & $100$ & $4$ & $8$ & $16$ & $50$ & $100$ & $4$ & $8$ & $16$ & $50$ & $100$ \\
\midrule

\multirow{4}{*}{Coloring} 
 & Success Rate     & 
 \gradientcell{0.83} & \gradientcell{0.94} & \gradientcell{0.91} & \gradientcell{0.92} & \gradientcell{0.98} & 

 \cellcolor{YellowGreen!30}{\gradientcell{1.0}} & 
 \cellcolor{YellowGreen!30}{\gradientcell{1.0}} & 
 \cellcolor{YellowGreen!30}{\gradientcell{0.96}} & 
 \cellcolor{YellowGreen!30}{\gradientcell{0.96}} & 
 \cellcolor{YellowGreen!30}{\gradientcell{0.97}} & 
 
 \gradientcell{0.8} & \gradientcell{1.0} & \gradientcell{0.77} & \gradientcell{0.93} & \gradientcell{0.81} & 
 \gradientcell{1.0} & \gradientcell{1.0} & \gradientcell{0.93} & - & - & 
 \gradientcell{1.0} & \gradientcell{0.86} & \gradientcell{1.0} & \gradientcell{0.98} & \gradientcell{0.98} & 
 - & - & - & - & - \\

& Rounds to Converge          & 
 8 & 8 & 8 & 11 & 15 & 
 \cellcolor{YellowGreen!30}{8} & \cellcolor{YellowGreen!30}{8} & \cellcolor{YellowGreen!30}{8} & \cellcolor{YellowGreen!30}{11} & \cellcolor{YellowGreen!30}{11} & 
 8 & 8 & 8 & 13 & 19 & 
 8 & 8 & 8 & >40 & >40 & 
 8 & 8 & 8 & 13 & 15 & 
 - & - & - & - & - \\
 
 & Token Cost ($\times 10^3$)            & 
 125 & 274 & 481 & 2,993 & 10,431 & 
 \cellcolor{YellowGreen!30}{109} & \cellcolor{YellowGreen!30}{233} & \cellcolor{YellowGreen!30}{508} & \cellcolor{YellowGreen!30}{2,906} & \cellcolor{YellowGreen!30}{5,699} & 
 116 & 270 & 555 & 4,362 & 18,108 & 
 95 & 205 & 444 & \$\$ & \$\$ & 
  104 & 204 & 420 & 3,115 & 8,102 & 
 - & - & - & - & - \\
 
 & Runtime (sec)               & 
 72 & 108 & 111 & 464 & 929 & 
 \cellcolor{YellowGreen!30}{66} & \cellcolor{YellowGreen!30}{108} & \cellcolor{YellowGreen!30}{117} & \cellcolor{YellowGreen!30}{378} & \cellcolor{YellowGreen!30}{652} & 
 118 & 120 & 158 & 507 & 1209 & 
 72 & 97 & 118 & - & - & 
 85 & 101 & 150 & 432 & 865 & 
 - & - & - & - & - \\
\midrule

\multirow{4}{*}{Matching} 
 & Success Rate     & 
 \gradientcell{0.5} & \gradientcell{0.75} & \gradientcell{0.88} & \gradientcell{0.68} & \gradientcell{0.68} & 
 \gradientcell{0.5} & \gradientcell{1.0} & \gradientcell{0.94} & \gradientcell{0.68} & \gradientcell{0.68} & 
 \gradientcell{1.0} & \gradientcell{0.75} & \gradientcell{1.0} & \gradientcell{0.64} & \gradientcell{0.0} & 
 \gradientcell{1.0} & \gradientcell{0.5} & \gradientcell{1.0} & - & - & 
 \cellcolor{YellowGreen!30}{\gradientcell{1.0}} & 
 \cellcolor{YellowGreen!30}{\gradientcell{0.88}} & 
 \cellcolor{YellowGreen!30}{\gradientcell{1.0}} & 
 \cellcolor{YellowGreen!30}{\gradientcell{0.80}} & 
 \cellcolor{YellowGreen!30}{\gradientcell{0.66}} & 
 
 \gradientcell{1.0} & \gradientcell{1.0} & \gradientcell{0.75} & \gradientcell{0.24} & \gradientcell{0.42} \\

 & Rounds to Converge          & 
 8 & 8 & 8 & 11 & 15 & 
 8 & 8 & 8 & 11 & 11 & 
 8 & 8 & 8 & 13 & 19 & 
 8 & 8 & 8 & >40 & >40 & 
 \cellcolor{YellowGreen!30}{8} & \cellcolor{YellowGreen!30}{8} & \cellcolor{YellowGreen!30}{8} & \cellcolor{YellowGreen!30}{13} & \cellcolor{YellowGreen!30}{15} & 
 8 & 8 & 8 & 3 & 3 \\
 
 & Token Cost ($\times 10^3$)            & 
 112 & 249 & 538 & 2,888 & 9,847 & 
 102 & 217 & 482 & 2,684 & 5,274 & 
 107 & 268 & 547 & 4,265 & 18,283 & 
 90 & 198 & 423 & \$\$ & \$\$ & 
 \cellcolor{YellowGreen!30}{94} & \cellcolor{YellowGreen!30}{203} & \cellcolor{YellowGreen!30}{395} & \cellcolor{YellowGreen!30}{2,951} & \cellcolor{YellowGreen!30}{7,827} & 
 107 & 340 & 829 & 1,124 & 3,016 \\

 & Runtime (sec)               & 
 111 & 123 & 141 & 311 & 910 & 
 88 & 95 & 120 & 306 & 577 & 
 77 & 81 & 109 & 462 & 1,400 &  
 67 & 90 & 124 & - & - & 
 \cellcolor{YellowGreen!30}{83} & \cellcolor{YellowGreen!30}{88} & \cellcolor{YellowGreen!30}{112} & \cellcolor{YellowGreen!30}{412} & \cellcolor{YellowGreen!30}{760} &
 74 & 134 & 196 & 182 & 621 \\
\midrule

\multirow{4}{*}{VertexCover} 
 & Success Rate     & 

 \cellcolor{YellowGreen!30}{\gradientcell{0.83}} & 
 \cellcolor{YellowGreen!30}{\gradientcell{0.79}} & 
 \cellcolor{YellowGreen!30}{\gradientcell{0.99}} & 
 \cellcolor{YellowGreen!30}{\gradientcell{0.88}} & 
 \cellcolor{YellowGreen!30}{\gradientcell{0.81}} & 
 
 \gradientcell{0.0} & \gradientcell{0.29} & \gradientcell{0.64} & \gradientcell{0.80} & \gradientcell{0.85} & 
 
 \gradientcell{1.0} & \gradientcell{0.0} & \gradientcell{0.95} & \gradientcell{0.84} & \gradientcell{0.80} & 
 
 \gradientcell{0.67} & \gradientcell{0.57} & \gradientcell{0.73} & - & - & 
 
 \gradientcell{0} & \gradientcell{0.57} & \gradientcell{0.3} & \gradientcell{0.57} & \gradientcell{0.72} & 
 
 - & - & - & - & - \\

 & Rounds to Converge          & 
 \cellcolor{YellowGreen!30}{8} & \cellcolor{YellowGreen!30}{8} & \cellcolor{YellowGreen!30}{8} & \cellcolor{YellowGreen!30}{11} & \cellcolor{YellowGreen!30}{15} & 
 8 & 8 & 8 & 11 & 11 & 
 8 & 8 & 8 & 13 & 18 & 
 8 & 8 & 8 & >40 & >40 & 
 8 & 8 & 8 & 13 & 15 & 
 - & - & - & - & - \\
 
 & Token Cost ($\times 10^3$)            & 
 \cellcolor{YellowGreen!30}{124} & \cellcolor{YellowGreen!30}{350} & \cellcolor{YellowGreen!30}{1,210} & \cellcolor{YellowGreen!30}{3,391} & \cellcolor{YellowGreen!30}{11,536} & 
 122 & 259 & 569 & 3,185 & 6,455 & 
 131 & 307 & 635 & 5,066 & 20,335 & 
 110 & 264 & 489 & \$\$ & \$\$ & 
 113 & 226 & 463 & 3,359 & 9,109 & 
 - & - & - & - & - \\
 
 & Runtime (sec)               & 
 \cellcolor{YellowGreen!30}{91} & \cellcolor{YellowGreen!30}{118} & \cellcolor{YellowGreen!30}{321} & \cellcolor{YellowGreen!30}{446} & \cellcolor{YellowGreen!30}{1,057} & 
 96 & 111 & 133 & 341 & 670 & 
 88 & 116 & 130 & 621 & 1,516 & 
 80 & 172 & 99 & - & - & 
 94 & 107 & 126 & 396 & 912 & 
 - & - & - & - & - \\
\midrule

\multirow{4}{*}{LeaderElection} 
 & Success     & 
 \textcolor{Mahogany}{\xmark} & 
 \textcolor{Mahogany}{\xmark} & 
 \textcolor{Mahogany}{\xmark} & 
 \textcolor{Mahogany}{\xmark} & 
 \textcolor{Mahogany}{\xmark} & 

 \textcolor{Mahogany}{\xmark} & 
 \textcolor{Mahogany}{\xmark} & 
 \textcolor{Mahogany}{\xmark} & 
 \textcolor{Mahogany}{\xmark} & 
 \textcolor{Mahogany}{\xmark} & 

 \textcolor{Mahogany}{\xmark} & 
 \cellcolor{YellowGreen!30}{\textcolor{ForestGreen}{\checkmark}} & 
 \cellcolor{YellowGreen!30}{\textcolor{ForestGreen}{\checkmark}} & 
 \cellcolor{YellowGreen!30}{\textcolor{ForestGreen}{\checkmark}} & 
 \cellcolor{YellowGreen!30}{\textcolor{ForestGreen}{\checkmark}} & 

 \textcolor{Mahogany}{\xmark} & 
 \textcolor{Mahogany}{\xmark} & 
 \textcolor{ForestGreen}{\checkmark} & 
 - & 
 - & 

 \textcolor{Mahogany}{\xmark} & 
 \textcolor{ForestGreen}{\checkmark} & 
 \textcolor{Mahogany}{\xmark} & 
 \textcolor{Mahogany}{\xmark} & 
 \textcolor{Mahogany}{\xmark} & 

 \textcolor{ForestGreen}{\checkmark} & 
 \textcolor{Mahogany}{\xmark} & 
 \textcolor{Mahogany}{\xmark} & 
 \textcolor{Mahogany}{\xmark} & 
 \textcolor{Mahogany}{\xmark} \\

 & Rounds to Converge          &
 3 & 7 & 7 & 11 & 15 & 
 5 & 7 & 9 & 11 & 11 & 
 5 & \cellcolor{YellowGreen!30}{5} & \cellcolor{YellowGreen!30}{9} & \cellcolor{YellowGreen!30}{13} & \cellcolor{YellowGreen!30}{19} & 
 7 & 15 & 31 & >40 & >40 & 
 5 & 7 & 9 & 13 & 15 & 
 3 & 3 & 3 & 3 & 3 \\
 
 & Token Cost ($\times 10^3$)            & 
 29 & 220 & 438 & 2,985 & 10,371 & 
 51 & 182 & 608 & 2,929 & 5,858 & 
 54 & \cellcolor{YellowGreen!30}{124} & \cellcolor{YellowGreen!30}{707} & \cellcolor{YellowGreen!30}{4,595} & \cellcolor{YellowGreen!30}{18,784} & 
 90 & 645 & 5,216 & \textdollar\textdollar
 & \textdollar\textdollar
 & 
 48 & 168 & 512 & 3,179 & 8,168 & 
 29 & 72 & 194 & 1,229 & 2,784 \\
 & Runtime (sec)               & 
 35 & 102 & 157 & 327 & 867 & 
 50 & 73 & 125 & 412 & 615 & 
 41 & \cellcolor{YellowGreen!30}{64} & \cellcolor{YellowGreen!30}{168} & \cellcolor{YellowGreen!30}{483} & \cellcolor{YellowGreen!30}{1,463} & 
 86 & 207 & 666 & - & - & 
 54 & 77 & 107 & 444 & 866 & 
 40 & 47 & 61 & 246 & 515 \\
\midrule

\multirow{4}{*}{Consensus} 
  & Success     & 
 \textcolor{ForestGreen}{\checkmark} & \textcolor{Mahogany}{\xmark} & \textcolor{ForestGreen}{\checkmark} & \textcolor{Mahogany}{\xmark} & \textcolor{Mahogany}{\xmark} & 
 \textcolor{ForestGreen}{\checkmark} & \textcolor{ForestGreen}{\checkmark} & \textcolor{ForestGreen}{\checkmark} & \textcolor{Mahogany}{\xmark} & \textcolor{Mahogany}{\xmark} & 
 \textcolor{ForestGreen}{\checkmark} & \textcolor{Mahogany}{\xmark} & \textcolor{ForestGreen}{\checkmark} & \textcolor{Mahogany}{\xmark} & \textcolor{Mahogany}{\xmark} & 
 \textcolor{Mahogany}{\xmark} & \textcolor{ForestGreen}{\checkmark} & \textcolor{Mahogany}{\xmark} & - & - & 

 \textcolor{ForestGreen}{\checkmark} & 
 \textcolor{Mahogany}{\xmark} & 
 \textcolor{Mahogany}{\xmark} & 
 \textcolor{Mahogany}{\xmark} & 
 \textcolor{Mahogany}{\xmark} & 
 
 \cellcolor{YellowGreen!30}{\textcolor{ForestGreen}{\checkmark}} &
\cellcolor{YellowGreen!30}{\textcolor{ForestGreen}{\checkmark}} &
\cellcolor{YellowGreen!30}{\textcolor{ForestGreen}{\checkmark}} &
\cellcolor{YellowGreen!30}{\textcolor{ForestGreen}{\checkmark}} &
\cellcolor{YellowGreen!30}{\textcolor{ForestGreen}{\checkmark}} \\

 & Rounds to Converge          & 
 3 & 7 & 7 & 11 & 15 & 
 5 & 7 & 9 & 11 & 11 & 
 5 & 5 & 9 & 13 & 19 & 
 7 & 15 & 31 & >40 & >40 & 
 5 & 7 & 9 & 13 & 15 & 
 \cellcolor{YellowGreen!30}{3} & 
 \cellcolor{YellowGreen!30}{3} & 
 \cellcolor{YellowGreen!30}{3} & 
 \cellcolor{YellowGreen!30}{3} & 
 \cellcolor{YellowGreen!30}{3}  \\
 
 & Token Cost ($\times 10^3$)            & 
 26 & 211 & 379 & 2,892 & 10,664 & 
 51 & 170 & 594 & 2,793 & 5,734 & 
 52 & 122 & 629 & 4,222 & 18,283 & 
 76 & 620 & 5,183 & \textdollar\textdollar & \textdollar\textdollar & 
 49 & 163 & 505 & 3,061 & 8,336 & 
 
 \cellcolor{YellowGreen!30}{24} & 
 \cellcolor{YellowGreen!30}{61} & 
 \cellcolor{YellowGreen!30}{175} & 
 \cellcolor{YellowGreen!30}{1,151} & 
 \cellcolor{YellowGreen!30}{2,611} \\
 
 & Runtime (sec)               & 
 22 & 101 & 86 & 378 & 1,010 & 
 54 & 77 & 134 & 372 & 733 & 
 48 & 62 & 145 & 472 & 1,400 & 
 71 & 201 & 526 & - & - & 
 60 & 116 & 162 & 402 & 959 & 
 \cellcolor{YellowGreen!30}{26} & 
 \cellcolor{YellowGreen!30}{32} & 
 \cellcolor{YellowGreen!30}{50} & 
 \cellcolor{YellowGreen!30}{228} & 
 \cellcolor{YellowGreen!30}{922} \\
\bottomrule
\end{tabular}
\vspace{-4mm}
\end{table*}

\subsection{\revtwo{R2O1}{Multi-Agent Evaluation}}
\label{subsec:multi_agent}

We evaluate how communication architecture affects coordination success,
scalability, and execution cost. Using the MAFBench coordination pipeline, we
fix the LLM, task semantics, prompts, stopping criteria, and metrics, and vary
only the topology through controlled graph structures. \revone{R1O5}{Agents
communicate only with direct neighbors, so what each observes follows from the
topology alone. A globally readable shared state would make every structure
equally dense and remove the variable under study.} This isolates the effect of
communication architecture as agent populations grow.

\subsubsection{Communication, Coordination, and Scaling}
\label{sec:Coordination_section}
We evaluate how communication topology shapes coordination outcomes under
constrained message passing using
AGENTSNET~\cite{coordinationcollaborativereasoning}. All experiments fix the
model backend, task definitions, prompts, execution protocol, and round budgets
through the unified MAFBench infrastructure, and vary only the interaction
structure (Section~\ref{subsec:bench_coordination}). We compare graph-based
topologies (small-world~\cite{watts1998collective}, scale-free~\cite{scale_free},
and Delaunay~\cite{delaunay}), role-based pipelines (sequential and
hierarchical), and a fully connected structure standing in for GABM
centralization. Network size varies over \(n \in \{4, 8, 16, 50, 100\}\). We
report task success, rounds to convergence, token usage, and
runtime (Table~\ref{tab:scalability-results}). Bounded tasks run a minimum of
eight rounds. Success is binary for the global agreement tasks
\emph{LeaderElection} and \emph{Consensus}.\footnote{Detailed configurations,
success definitions for all tasks, and metric formulations are provided in the
\uline{\textcolor{blue}{\href{https://github.com/CoDS-GCS/MAFBench/blob/main/supplementary_materials/APPENDICIES.pdf}{supplementary materials}}}.}
The maximum follows \(T = 2D + 1\), where \(D\) is the graph diameter, so
information has enough rounds to propagate.\footnote{The graph diameter \(D\) is
the maximum shortest-path distance between any pair of nodes in the
communication graph.}
\revone{R1O5}{Under direct message passing, connectivity and visibility coincide,
so full connectivity gives global visibility, and is the only structure reaching
consensus. A shared environment decouples the two, because a structurally sparse
star may still expose global state to every agent. 
Coordination under shared-state environments is future work.
}

Local tasks tolerate sparse structure. \emph{Coloring} succeeds across every
graph family, and \emph{Matching} holds from $0.5$ for all graph-based
families before $n{=}50$, because each agent needs only its neighbors. Topology
dependence is stronger on tasks that need global information.
\emph{VertexCover} minimizes a global objective from local decisions, and
scale-free graphs dominate at large scale in both success and efficiency. For
\emph{LeaderElection}, small-world and scale-free structures fail at every size,
and Delaunay graphs succeed only at moderate and large scale at high
coordination cost. Role-based hierarchies and centralized interaction succeed
only at very small scale, then degrade rapidly. For \emph{Consensus}, the fully
connected topology is the only structure that succeeds consistently, converging
in three rounds independent of network size. Every graph-based topology fails
despite substantially higher runtime and token expenditure. Consensus requires
every agent to receive information from every other agent within bounded rounds,
and sparse graphs create fragmented clusters that cannot merge. This answers
RQ4.\footnote{Consensus dynamics across all topologies and network sizes are
visualized in the
\uline{\textcolor{blue}{\href{https://github.com/CoDS-GCS/MAFBench/blob/main/supplementary_materials/APPENDICIES.pdf}{supplementary materials}}},
showing persistent fragmentation in sparse graphs.}

Coordination scalability is therefore an architectural property of communication
structure. Local coordination benefits from sparse topologies that preserve
neighborhood structure. No single topology performs well across all tasks, so
communication architecture should match the information flow the task requires.
\revtwo{R2O3}{Topologies are instantiated by MAFBench's rewriting engine rather
than by vendor frameworks, so these results are structure-level and carry no
framework-implementation confound.}

\vspace{-2mm}
\subsection{Scope and Reflection}
\label{subsec:threats}
Each experiment fixes the LLM and task to isolate one architectural dimension,
so results are not verified across all model families, scales, or workloads. We
partially address this with multiple LLM backends. \revone{R1O2}{Concordia is
one GABM implementation, so its absolute numbers do not characterize the
paradigm. The minimal skeleton supplies the paradigm-level reference point.}
\revthree{R3W3}{LLM outputs vary across identical runs, so we repeat the
planning-interface experiment on MATH-100 five times each for GPT-5.6-Terra,
GPT-5.6-Luna, GPT-OSS-20B, and Llama-3.1-8B, spanning the capability
range.\footnote{Per-model means and population standard deviations are in the
\uline{\textcolor{blue}{\href{https://github.com/CoDS-GCS/MAFBench/blob/main/supplementary_materials/APPENDICIES.pdf}{supplementary materials}}}.}
Accuracy varies by at most $4.1$ points and schema-violation rate by at most
$3.3$, small relative to the differences the paper attributes to the interface,
so single-run reporting preserves the architectural rankings. Aggregation
absorbs further noise, because every score is a mean over 100 questions, and
memory and coordination scores average over subtasks and graph instances too.}
\revthree{R3M1}{All five research questions are
answered} (Section~\ref{sec:hypotheses}). Retrieval dominates multi-hop
retrieval and in-session learning, long-range understanding benefits from both
retrieval architecture and model capability, and knowledge revision fails
\revthree{R3W4}{across all evaluated framework-native architectures and both
LLMs} (RQ1). \revthree{R3W1}{Schema-constrained planning reduces accuracy
through formatting failures below the frontier, above which the plan itself is
the cost} (RQ2). \revone{R1O4}{Role labels do not activate domain reasoning, and
bound capability helps only when it supplies procedure} (RQ3). Sparse topologies
fail on global agreement (RQ4). \revone{R1O2}{Environment mediation costs
milliseconds, so Concordia's order-of-magnitude overhead is implementation
cost} (RQ5). These effects hold across the \textit{evaluated} frameworks,
datasets, and LLMs, converging on one theme: architectural design, not model
intelligence alone, governs multi-agent behavior.
  \section{Multi-Agent Design Principles}
\label{sec:design_principles}

This section distills results from Section~\ref{sec:eval} into actionable design
principles for multi-agent LLM systems. Each principle is grounded in controlled
empirical evidence and states both where it applies and where it does not.

\smallskip
\noindent\textbf{Principle 1: Orchestration overhead is the dominant scalability constraint.}
\emph{\revone{R1O2}{Judge orchestration by its execution structure, not by its paradigm.}}
Framework execution structure governs latency and throughput even for trivial
workloads (Section~\ref{subsec:overhead}, \revthree{R3M1}{RQ5}). Direct LLM
calls incur minimal cost. \revone{R1O2}{Graph- and role-based frameworks add a
thin wrapper over the model call, with Agno the exception. Concordia's GABM
execution produces orders-of-magnitude higher latency from game-master
mediation, persistent loops, and state serialization. A minimal implementation
of the same paradigm does not.} These effects arise from orchestration
semantics, not task complexity.
\emph{Scope:} this principle applies to latency-sensitive and
throughput-sensitive production systems such as high-volume document processing.
For tasks that require multi-round deliberation such as social simulation,
deeper orchestration may be justified despite overhead.
\revone{R1O2}{Measure the framework rather than assuming its paradigm's cost.
Two implementations of one paradigm differed by three orders of magnitude here.}\revtwo{R2O4}{}
The key decision is whether the task requires coordination across rounds or
decomposes into independent single-agent calls.\shorten

\smallskip
\noindent\textbf{Principle 2: \revthree{R3W4}{Framework-native} memory must be architected around task semantics, not context size.}
\emph{Choose memory architecture from task needs, and never rely on larger context windows.}
Architectural design governs memory behavior rather than available
context (Section~\ref{subsec:memoryagentbench}, \revthree{R3M1}{RQ1}).
Retrieval-first architectures dominate factual recall and long-range reasoning
by isolating relevant evidence. Accumulation-based memory degrades as unfiltered
history expands, even under large context budgets. Hybrid designs achieve the
strongest in-session learning by filtering accumulated signals through
retrieval. No single architecture performs well across all competencies.
Selective Forgetting fails \revthree{R3W4}{across all evaluated framework-native
architectures}, which treat memory as append-only and lack explicit editing
mechanisms.
\emph{Scope:} use retrieval-first memory for knowledge-intensive tasks such as
multi-hop QA. Use bounded accumulation for session-oriented tasks such as
tutoring. Never replace architectural memory control with larger context
windows. Applications requiring controlled knowledge revision, such as
correcting outdated facts, need \revthree{R3W4}{primitives no evaluated
framework provides natively}.

\smallskip
\noindent\textbf{Principle 3: Planning \revone{R1O3}{interfaces} should be permissive, and \revthree{R3W1}{planning itself optional}.}
\emph{Elicit a plan only when the model needs one, and keep the interface permissive.}
\revone{R1O3}{The interface and the plan carry separate
costs} (Section~\ref{subsec:planning_results}, \revthree{R3M1}{RQ2}).
Formatting failures, not reasoning errors, dominate accuracy loss for weaker
models, and free-form planning removes them at modest runtime cost.
\revthree{R3W1}{Frontier models never violate the schema, so that cost vanishes.
The plan still does not pay. Accuracy remains almost flat under either
interface, and every model pays the extra call.}
\emph{Scope:} \revthree{R3W1}{skip the plan where the task fits the model's
context and it plans internally.} Enforce structure only when downstream
components require machine-readable output, because recovering from a parse
failure costs more than post-processing a loose plan.

\smallskip
\noindent\textbf{Principle 4: Specialization should be procedural, and role labels secondary.}
\emph{Encode expertise as workflows, not as role names.}
Assigning professional roles does not activate domain-specific
reasoning (Section~\ref{sec:specialization}, \revthree{R3M1}{RQ3}), because
labels do not constrain how the model processes data. Self-generated planning
stabilizes behavior without accuracy gains. Expert-guided conditioning delivers
large improvements by imposing explicit workflows. \revone{R1O4}{Binding a
capability is not sufficient either. A dataset-profile tool grants data access
without procedure, and helps only where the obstacle is knowing the data.}
\emph{Scope:} implement specialization through reusable procedural templates,
such as data cleaning pipelines. \revone{R1O4}{Supply tools when the agent lacks
information and procedure when it lacks method, because the two are not
interchangeable.} Role labels may still support routing or delegation. These
findings use GPT-4o-mini across all five CatDB datasets, but the effect may differ for models with stronger instruction following.

\smallskip
\noindent\textbf{Principle 5: Coordination is governed by communication topology, not by agent intelligence.}
\emph{Design communication structure explicitly, because more interaction cannot fix bad topology.}
Communication topology rather than interaction budget determines coordination
success and scalability (Section~\ref{sec:Coordination_section},
\revthree{R3M1}{RQ4}). For local coordination tasks, scale-free graphs offer the
best success-cost trade-offs through efficient information routing. Sparse
graphs with long paths fail on global agreement even under large round and token
budgets. Only fully connected topology achieves consistent consensus, because
every agent must receive information from every other agent within bounded
rounds.
\emph{Scope:} full connectivity enables global agreement but incurs quadratic
communication cost. For local coordination tasks such as distributed conflict
resolution, where agents need only neighborhood information, sparse topologies
provide better cost-success trade-offs. For tasks requiring system-wide
consensus such as leader election, fully connected or near-complete topologies
are necessary. Select communication architecture from whether the task requires
local or global information flow.\shorten

\smallskip
\noindent\textbf{Principle 6: System interfaces dominate multi-agent behavior.}
\emph{Engineer robustness and scalability through interfaces, not prompts.}
Across all five dimensions, performance limits arise from framework execution
semantics and interface design rather than task difficulty or model capability.
Orchestration determines latency and throughput, memory architecture controls
recall and scalability, planning interfaces govern accuracy, specialization
depends on procedural structure, and topology determines coordination success.
\emph{Scope:} invest engineering effort in interface design first. Select
orchestration depth, memory access mode, planning format, and communication
topology before tuning prompts or selecting models. Architectural bottlenecks
persist across model generations, and additional rounds, tokens, or prompt
refinement cannot compensate for them.
  \section{Future Directions}
\label{sec:future_work}

This section outlines research directions motivated by the architectural
limitations identified in our evaluation. We identify what is missing and state
the open research question.

\smallskip
\noindent\textbf{Extending MAFBench.}
MAFBench currently evaluates five architectural dimensions under single-model
configurations. Three extensions would broaden its scope. First, built-in
cross-model sweeps with multi-run statistical reporting would evaluate every
dimension across multiple LLM backends and strengthen reproducibility. Second,
industry-representative workload profiles, such as long-running sessions,
high-concurrency pipelines, and mixed-task workflows, would complement the
current academic benchmarks. Third, \revthree{R3Q3}{controlled tool-use
benchmarks would isolate framework-level tool orchestration from model-level
tool calling once frameworks expose architectural control over tool selection.}
Repeated LLM-based evaluation at scale remains a practical barrier that requires
cost-aware scheduling and selective replication.

\smallskip
\noindent\textbf{Memory Editing and Lifecycle Management.}
\revthree{R3W4}{Every evaluated framework-native architecture treats memory as
append-only} (Section~\ref{subsec:memoryagentbench}), providing no controlled
revision or dependency-aware deletion. \revthree{R3W4}{Adaptive memory systems
and plugin layers~\cite{kang2025memory,xu2025mem} update and refine stored
knowledge, but lack mechanisms for safe removal.} Future memory architectures
should support explicit state updates, namely revise, remove, and invalidate,
with dependency tracking across stored facts. A concrete solution requires
versioned fact stores with multi-version concurrency control, write-ahead
logging for reversible updates, and conflict resolution for concurrent agent
writes. Memory stores in multi-agent systems therefore need the transactional
semantics, versioning, and consistency guarantees that databases already
provide.\shorten

\smallskip
\noindent\textbf{Multi-Agent Frameworks for Data-Intensive Applications.}
Multi-agent systems increasingly serve data-intensive applications, such as
knowledge graph question answering, ETL pipeline orchestration, data
integration, and conversational analytics. The trade-offs exposed by MAFBench
directly affect these workloads. Memory scalability governs how agents manage
large knowledge bases, coordination topology determines how agents distribute
data processing, and orchestration overhead constrains real-time pipelines.
Current practice requires manual architecture selection per application, which
does not scale. The long-term goal is agentic systems that autonomously
construct multi-agent architectures to automate complex industry roles, such as
data engineering and financial analysis. This treats multi-agent system
construction as a compilation problem that translates high-level task
specifications into concrete orchestration, memory, planning, specialization,
and coordination configurations. Section~\ref{sec:design_principles} supplies
the optimization objectives and MAFBench the evaluation infrastructure, so
connecting them into an automated pipeline is an open systems problem at the
intersection of multi-agent design and data management.\shorten

\smallskip
\noindent\textbf{Safety and Security.}
Our taxonomy identifies three dimensions with direct safety implications.
Communication structure $\mathcal{C}$ affects adversarial information
propagation. Storage design $\mathcal{S}$ influences persistence of sensitive
data in agent memory. Orchestration $\mathcal{O}$ governs execution of external
actions without authorization. Future benchmarks should measure concrete
security properties under different configurations: prompt injection propagation
rate, memory poisoning persistence, and unauthorized tool execution rate.

  \vspace{-1mm}
\section{Related Work}

Existing benchmarks focus on evaluating individual model or agent capabilities, but do not support controlled experimental analysis of the system-level factors that govern multi-agent framework performance. 
Works such as HELM~\cite{helm} and Chatbot Arena~\cite{chiang2024chatbot} compare language models under standardized evaluation protocols, ensuring that observed differences are attributable to the model while treating the execution framework as fixed. 
This paradigm extends to agent settings in benchmarks such as AgentBench~\cite{agentbench}, WebArena~\cite{webarena}, and SWE-bench~\cite{swebench}, which evaluate agents in realistic and interactive environments but keep the environment, execution pipeline, and framework design unchanged across comparisons. 
As a result, these benchmarks measure capability improvements without enabling analysis of which architectural factors dominate performance or how design choices impact system behavior.

A second line of work explores multi-agent architectures more directly, but lacks controlled isolation of framework-level variables. 
BOLAA~\cite{bolaa} evaluates multiple agent architectures combined with different LLM backbones, jointly varying both factors and confounding their effects. 
MetaGPT~\cite{metagpt} proposes a structured multi-agent pipeline based on standardized operating procedures, but evaluates a single fixed design without varying architectural dimensions. 
Consequently, prior work does not provide controlled experiments to identify the principles governing multi-agent system design or to quantify the impact of individual architectural decisions. 
MAFBench addresses this gap by fixing the underlying LLM and task while varying architectural dimensions, enabling controlled attribution of performance differences to framework design choices such as memory, planning interfaces, and communication topology.
  \vspace{-1mm}
\section{Conclusion}
\label{sec:conclusion}
We introduce an architectural taxonomy across five dimensions (orchestration,
memory, planning, specialization, and communication topology),
\revthree{R3M1}{yielding five research questions, all answered}. MAFBench
enables controlled, architecture-level comparison under fixed models and tasks.
Our study across nine \revone{R1O2}{evaluated} frameworks shows that
architectural design, not model intelligence alone, governs performance.
\revone{R1O3}{Rigid planning schemas make up to 84.7\% of questions unparseable,
so formatting failures rather than reasoning errors dominate accuracy loss.}
Only fully connected topology achieves consensus at scale.
\revone{R1O4}{Role labels activate no domain reasoning, but bound procedure
lifts F1 above 94.} \revone{R1O2}{Environment mediation costs milliseconds in a
minimal implementation, isolating a production framework's order-of-magnitude
overhead as implementation cost.} Memory architecture governs recall and
scalability independent of context window size, and \revthree{R3W4}{no evaluated
framework natively supports} knowledge revision. These findings parallel the
trade-offs that drove database systems toward modular designs, motivating
research on adaptive architectures, principled coordination, and automated
compilation of agentic systems.\shorten

  \balance
  \bibliographystyle{ACM-Reference-Format}
  \bibliography{All_Bib}
\end{document}